\documentclass[10pt,twocolumn,letterpaper]{article}

\usepackage{wacv}              

\usepackage{times}
\usepackage{epsfig}
\usepackage{graphicx}
\usepackage{amsmath}
\usepackage{amssymb}

\usepackage{booktabs}
\usepackage{wrapfig}

\newtheorem{lemma}{Lemma}

\usepackage{algorithm}
\usepackage{algpseudocode}
\usepackage[accsupp]{axessibility}
\newcommand\blfootnote[1]{%
  \begingroup
  \renewcommand\thefootnote{}\footnote{#1}%
  \addtocounter{footnote}{-1}%
  \endgroup
}

\newcommand{\cD}{\mathcal{D}}

\newcommand{\cA}{\mathcal{A}}

\newcommand{\cH}{\mathcal{H}}
\newcommand{\cY}{\mathcal{Y}}

\usepackage[breaklinks=true,bookmarks=false]{hyperref}
\usepackage[capitalize]{cleveref}
\crefname{section}{Sec.}{Secs.}
\Crefname{section}{Section}{Sections}
\Crefname{table}{Table}{Tables}
\crefname{table}{Tab.}{Tabs.}

\def\wacvPaperID{2141} 
\def\confName{WACV}
\def\confYear{2024}

\begin{document}

\title{Meta-Learned Attribute Self-Interaction Network for Continual and Generalized\\ Zero-Shot Learning}
\author{Vinay Verma$^{2\star}$, Nikhil Mehta$^{2\star}$, Kevin J Liang$^3$, Aakansha Mishra$^4$, Lawrence Carin$^5$\\
$^2$Duke University, $^3$FAIR, Meta, $^4$IITG, $^5$KAUST\\
{\tt\small \{$^1$vverma.vinayy,$^2$nikhilmehta.dce\}@gmail.com}
}

\maketitle

\begin{abstract}
		Zero-shot learning (ZSL) is a promising approach to generalizing a model to categories unseen during training by leveraging class attributes, but challenges remain. Recently, methods using generative models to combat bias towards classes seen during training have pushed state of the art, but these generative models can be slow or computationally expensive to train. Also, these generative models assume that the attribute vector of each unseen class is available a priori at training, which is not always practical. Additionally, while many previous ZSL methods assume a one-time adaptation to unseen classes, in reality, the world is always changing, necessitating a constant adjustment of deployed models. Models unprepared to handle a sequential stream of data are likely to experience catastrophic forgetting. We propose a Meta-learned Attribute self-Interaction Network (MAIN) for continual ZSL. By pairing attribute self-interaction trained using meta-learning with inverse regularization of the attribute encoder, we are able to outperform state-of-the-art results without leveraging the unseen class attributes while also being able to train our models substantially faster ($>100\times$) than expensive generative-based approaches. We demonstrate this with experiments on five standard ZSL datasets (CUB, aPY, AWA1, AWA2, and SUN) in the generalized zero-shot learning and continual (fixed/dynamic) zero-shot learning settings. Extensive ablations and analyses demonstrate the efficacy of various components proposed.
	\end{abstract}
	\vspace{-2em}
	\section{Introduction} \blfootnote{$^\ast$Equal contribution.}
	\label{sec:intro}
	Deep learning has demonstrated the ability to learn powerful models given a sufficiently large, labeled dataset of a pre-defined set of classes~\cite{krizhevsky2012imagenet,he2016deep}.
	However, such models often generalize poorly to classes unseen during training.
	In an ever-evolving world in which new concepts or applications are to be expected, this brittleness can be an undesirable characteristic. In recent years, zero-shot learning (ZSL) \cite{akata2013label,verma2019meta,skorokhodov2021class,han2020learning} has been proposed as an alternative framework to classify the novel class data.
	ZSL approaches seek to leverage auxiliary information about these new classes, often in the form of class attributes. This side information allows for reasoning about the relations between classes, enabling the adaptation of the model to recognize samples from one of the novel classes. In the more general setting, a ZSL model should be capable of classifying inputs from both seen and unseen classes; this difficult setting is commonly referred to as generalized zero-shot learning (GZSL)~\cite{xian2018zero,vermageneralized}.
	
	Some of the strongest results in GZSL~\cite{vermageneralized,schonfeld2019generalized,vaegan,han2020learning,verma2019meta,kumar2019generative} have come from utilizing generative models. 
	By learning a generative mapping between the attributes and the data, synthetic samples can be conditionally generated from the unseen class attributes.
	The model can then be learned in the usual supervised manner on the joint set of seen and (generated) unseen class data, mitigating model bias toward the seen classes.
	While effective, training the requisite generative model, generating data, and training the model on this combined dataset can be expensive~\cite{narayan2020latent,li2019rethinking,verma2019meta}. Furthermore, these generative methods require the unseen class attributes during training; if the unseen classes are not yet known, defining attributes for unseen classes is challenging and error-prone. Thus, the requirement of knowing the unseen class attributes during training makes these methods less practical, hindering their applicability to the real world.
	
	If assuming a one-time adaptation from a pre-determined set of training classes, the cost of such a one-off process may be considered acceptable, but it may present challenges if model adaptations need to be done repeatedly. Most ZSL methods commonly consider only one adaptation, but in reality environments are often dynamic, and new class data may appear sequentially. For example, if it is important for a model to be able to classify a previously unseen class, it is natural that a future data collection effort may later make labeled data from these classes available~\cite{skorokhodov2021class}. Alternatively, changing requirements may require the model to learn from and then generalize to entirely new seen and unseen classes~\cite{gautam2020generalized}. In such cases, the model should be able to learn from new datasets without catastrophically forgetting~\cite{mccloskey1989catastrophic} previously seen data, even if that older data is no longer available in its entirety. Thus, it is important that ZSL methods can work in continual learning settings as well.
	
	To address these issues, we propose the Meta-learned Attribute self-Interaction Network (MAIN) for Generalized Zero-Shot Learning (GZSL), which neither requires unseen class attributes \emph{a priori} nor any expensive model adaptions. The MAIN framework learns a unified visual embedding space for the class-specific attributes and the corresponding images belonging to a class. To extract the attribute embeddings, MAIN learns an attribute encoder that maps the semantic information in attributes to the visual embedding space. The attribute encoder in MAIN uses a novel self-interaction module for attributes and meta-learning to generalize the encoder to unseen class attributes. Moreover, MAIN incorporates a theoretically motivated inverse regularization loss, which preserves the semantic attribute information in the visual embedding space. In experiments, we discuss the importance of various components introduced as part of the MAIN framework. We show that training in MAIN is $100\times$ faster than the prior methods that use generative models. We extend MAIN to the continual GZSL setting using a small reservoir of samples and show that MAIN outperforms the recent alternatives proposed for CZSL. Extensive experiments on CUB, aPY, AWA1, AWA2, and SUN datasets demonstrate that MAIN achieves state-of-the-art results in ZSL, GZSL, and continual GZSL settings.
\vspace{-0.6em}
\section{Related Work}
\subsection{Zero-Shot Learning}
	
	The ZSL literature is vast, with approaches that can be roughly divided into ($i$) non-generative and ($ii$) generative approaches. Initial work \cite{akata2013label,akata2015evaluation,norouzi2013zero,hwang2014unified,fu2015zero,xian2016latent} mainly focused on non-generative methods, with a primary objective of learning a function from the seen classes that can measure the similarity between the image embeddings (typically features extracted from a pre-trained model) and the attribute embeddings. \cite{norouzi2013zero,lampert2014attribute,xian2016latent} measure the linear compatibility between the image and attribute embeddings; however, a linear map assumption does not hold for complex relationships between the two spaces. Another set of works~\cite{saligram2016learningJoint,kodirov2015unsupervisedDA} focuses on modeling the relation with bi-linear functions. These approaches show promising results for the ZSL setup (where only unseen classes are evaluated), but they perform poorly in the GZSL setting (where during inference, both seen and unseen classes are present).  

    Of late, generative approaches have been popular for GZSL. Due to rapid progress in generative modeling ($e.g.$, VAEs~\cite{kingma2014vae}, GANs~\cite{GAN}) generative approaches have been able to synthesize increasingly high-quality samples. For example, \cite{vermageneralized,xian2018feature,mishra2017generative,felix2018multi,schonfeld2019generalized,chou2021adaptive,vaegan,keshari2020generalized,shen2020invertible,narayan2020latent} have used conditional VAEs or GANs to generate samples for unseen classes conditioned on the class attributes, which can then be used for training alongside samples from the seen classes. Given the ability to generate as many samples as needed, these approaches can easily handle model bias towards seen classes, leading to promising results for both ZSL and GZSL~\cite{vaegan,han2020learning}. However, these generative models assume all seen, and unseen class attributes are present during training. In many practical settings, this can be a strong assumption, as the model may not know what the unseen classes will be ahead of time. Also, the full pipeline required learning a generative model, synthesizing samples for the unseen classes, and training a classifier on the given and synthesized data makes the generative model-based ZSL framework expensive.
	
	The aforementioned problems with generative models and recent promising results of non-generative models~\cite{skorokhodov2021class,liu2021isometric} motivate us to revisit non-generative approaches. \cite{skorokhodov2021class} propose an inexpensive approach for the GZSL and shows that normalization and initialization may play a crucial role for GZSL and outperform the expensive generative model. These normalization techniques can be tricky: high sensitivity to hyperparameters means small changes quickly degrade model performance. We propose a self-gating mechanism, meta-learning-based training, and multi-modal regularization in a non-generative framework. Even without tricky and unstable normalization, we can outperform the recent generative or normalization-based models by a significant margin.
		\vspace{-0.6em}
	\subsection{Zero-Shot Continual Learning}
	One of the desiderata of continual learning techniques is the forward transfer of previous knowledge to future tasks, which may not be known ahead of time; similarly, GZSL approaches seek to adapt models to the novel, unseen classes while still being able to classify the seen classes.
	As such, there are clear connections between the two problem settings.
	Some recent works~\cite{lopez2017gradient,kuchibhotla2022unseen,wei2020lifelong,skorokhodov2021class,gautam2020generalized,gautam2021generative} have drawn increasing attention towards continual zero-shot learning (CZSL). For example, \cite{wei2020lifelong} considers a task incremental learning setting, where task ID for each sample is provided during train and test, leading to an easier and perhaps less realistic setting than class incremental learning.
	A-GEM~\cite{chaudhry2018efficient} proposed a regularization-based model to overcome catastrophic forgetting while maximizing the forward transfer. 
	\cite{skorokhodov2021class} proposed a simple class normalization as an efficient solution to ZSL and extended it to CZSL, where the attributes for the unseen classes of future tasks are known a priori. This setting is referred to as Fixed Continual GZSL. Meanwhile, \cite{gautam2020generalized,gautam2021generative,kuchibhotla2022unseen} proposed a replay-based approach, showing state-of-the-art results in a more realistic setting for CSZL called Dynamic Continual GZSL, where the attributes for unseen classes of future tasks are not known a priori.
	
	\begin{figure}[!t]
		\centering	
		\includegraphics[width=1\columnwidth]{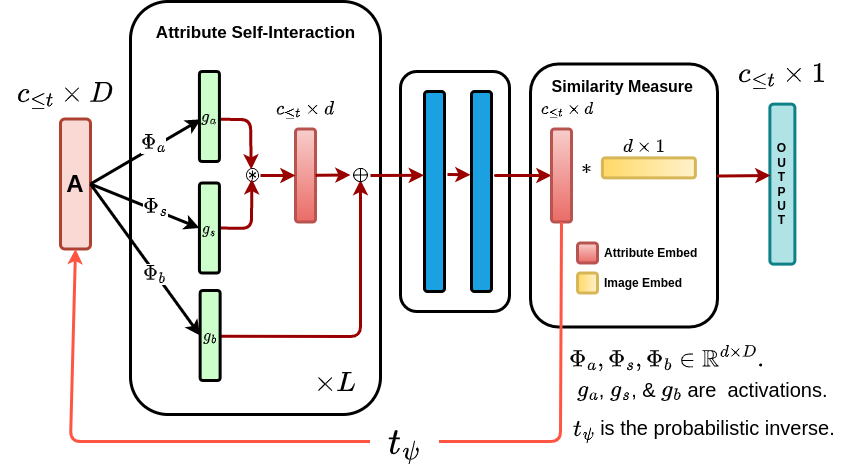}
		\vspace{-7mm}
		\caption{Meta-learned Attribute Self-Interaction Network.}
		\label{fig:czsl_proposed_arch}
		\vspace{-7mm}
	\end{figure}

\section{Proposed Approach}
\vspace{-3mm}
Our proposed Meta-Learned Attribute self-Interaction Network (MAIN)\footnote{\url{https://anonymous.4open.science/r/main\_zsl-C1BD}} for continual ZSL can be divided into three major components: ($i$) an attribute encoder designed using our novel self-interaction module, which maps the structured class-specific attribute to an attribute embedding in the visual feature space, ($ii$) Inverse Regularization (IR) that preserves the semantic information in the visual feature space, preventing over-fitting to the seen classes, and ($iii$) a meta-learning~\cite{nichol2018first} framework for training the attribute encoder with reservoir sampling to prevent catastrophic forgetting of previous tasks, while also learning a model that generalizes to novel classes without fine-tuning.

\subsection{Background and Notation}
\label{sec:background-notation}
The ZSL and GZSL follow the standard setting proposed by ~\cite{xian2018zero}, we provide more details in the supplementary material. Here we mostly focus on the continual setting.
Let $T_t=\{\cD_{tr}^t,\cD_{ts}^t\}$ be a task arriving at time $t$, where $\cD_{tr}^t$ and $\cD_{ts}^t$ are the train and test sets associated with the $t^{\mathrm{th}}$ task, respectively. In a continual learning setup, we assume that a set of tasks arrive sequentially, such that the training data for only the current task is made available. Let this sequence of tasks be $\{T_1,T_2,\dots T_K\}$, where at time $t$, the training data for only the $t^{\mathrm{th}}$ task is available. In continual learning, the goal is to learn a new task while preventing catastrophic forgetting of previously seen tasks. Hence, during testing, the model is evaluated on the current and previous tasks that the model has encountered before. We assume that for a given task $t$, we have $c^s_t$ number of seen classes that we can use for training and $c^u_t$ unseen/novel classes that we are interested in adapting our model to. Analogous to GZSL, during test time, we assume that samples come from any of the seen or unseen classes of the current or previous tasks, $i.e.$, the test samples for task $T_t$ contain $c_{\leq t}^s$ ($c_{\leq t}^s=\sum_{k=1}^tc_{k}^s$) number of seen classes and $c_{\leq t}^u$ number of unseen classes, where $c_{\leq t}^u$ will depend on the chosen evaluation protocol as described in supplementary. We denote $\cD_{tr}^t=\{x_i^t,y_i^t,a_{y_i},t\}_{i=1}^{N_{t}^s}$ to be the training data, where
$x_i^t\in \mathbb{R}^d$ is the visual feature of sample $i$ of task $t$, $y_i^t$ is the label of $x_i^t$, $a_{y_i}$ is the attribute/description vector of the label, $t$ is the task identifier (id) and $\cY_c$ is the label set for $c$ classes. Similarly, testing data for task $t$ is 
\begin{equation}
\cD_{ts}^t=\left\{\{x_i^t\}_{i=1}^{N_{t}^u},y_{x_i^t},\cA_t: \cA_t=\{a_y\}_{y=1}^{y=c_{\leq t}^s+c_{t}^u}\right\}
\end{equation}
where $y_{x_i^t}\in \{k \in \mathbb{N} \, | \, k \leq c_{\leq t}^s+c_{t}^u
\}$. For each task $T_t$, the seen and unseen classes are disjoint:  $\cY_{c_{t}^s}\cap \cY_{c_{t}^u}=\emptyset$. In each task $t$, seen and unseen classes have an associated attribute/description vector $a_y\in \cA_t$ that helps to transfer knowledge from seen to unseen classes. We will drop the class-specific indexing for attributes in $\{a_y\}$ and simply denote the attributes as $\{a\}$ in the following sections of the paper. We use $N_{t}^s$ and $N_{t}^u$ to denote the number of training and testing samples for the task $t$, respectively. We assume that there are a total of $S$ seen classes and $U$ unseen classes for each dataset, and the attribute set is $\cA\in\mathbb{R}^{S\cup U\times D}$, where $D$ is the attribute dimension. In the continual learning setup, test samples contain both unseen and seen classes up to task $t$, and the objective of continual ZSL is to predict future task classes while also overcoming catastrophic forgetting of the previous tasks.

\subsection{Self-Interaction for Attributes (SIA)}
\label{sec:self-interaction}

We develop a self-interaction module for the attribute vector  that can continually learn from sequential data and provide a robust class-representative vector in the visual embedding space. Let $\cA_{\leq t}^s \subset \mathbb{R}^D$ be the set of seen class attributes up to the training of task $T_t$, where  $|\cA_{\leq t}^s| = c_{\leq t}^s$. The self-interaction module comprises of three transformations parameterized as $\Phi_a$, $\Phi_s$, and $\Phi_b$ followed by activations $g_a$, $g_s$ and $g_b$, respectively. Consider an attribute encoder $f_\Phi$ with $L$ self-interaction modules, where the output of each module is denoted as $\mathbf{a}_{\ell}$. Given an attribute vector $\mathbf{a} \in \cA_{\leq t}^s$, we define $\mathbf{a}_{\ell+1}$ as:
\vspace{-0.5em}
\begin{align}
\small
\label{eq:self-interaction}
\mathbf{a}_{\ell + 1} = g_a\left(\Phi_a\left(\mathbf{a}_\ell\right)\right)*g_s\left(\Phi_s\left(\mathbf{a}_\ell\right)\right) + g_b\left(\Phi_b\left(\mathbf{a}_\ell\right)\right),
\end{align}
where $\mathbf{a}_0 = \mathbf{a}$. The final output of the attribute encoder $\mathbf{z} = f_\Phi(\mathbf{a})$ is computed by passing $\mathbf{a}_L$ to two fully-connected layers as shown in figure~\ref{fig:czsl_proposed_arch}. Next, we analyze two types of self-interactions, depending on type of activation functions.
\vspace{-7mm}
\subsubsection{Polynomial Kernels}
\vspace{-0.5em}
The self-interaction module described above is equivalent to the polynomial kernel when the activations $g_a$, $g_s$ and $g_b$ are linear. If we stack $L$ self-interaction modules, the output of layer-$\ell$ approximates the class of polynomials that grow exponentially in degree.
{\begin{lemma}[Polynomial Approximation]
Consider a model with $L$ layers of self-interaction modules as defined in (\ref{eq:self-interaction}) with parameters $\{\Phi^\ell_a, \, \Phi^\ell_s, \, \Phi^\ell_b\}_{\ell=1}^L$ and identity activation $g^{\ell}_a(x) = g^{\ell}_s(x) = g^{\ell}_b(x) = x$. Let input to the model be:  $\mathbf{a} = [ a_{1}, a_{3}, \dots,  a_{D} ]$. Then, the output of the model $\mathbf{a}_L$ approximates following class of polynomial functions:
\vspace{-0.7em}
\begin{small}\begin{align}\nonumber
    \Bigg\{ P_{\ell}\left(\mathbf{a}\right) = \sum_\beta w_\beta \, a_{1}^{\beta_1} \, a_{2}^{\beta_2} \dots a_{D}^{\beta_D} \enspace \Bigg| \enspace 0 \leq |\beta| \leq 2^{\ell}, \beta \in \mathbb{N}^D \Bigg\},
\end{align}\end{small}
where the sum is across multiple terms (monomials), $\beta = [\beta_1,\dots,\beta_D]$ is a vector containing the exponents of each attribute in a given term having degree $|\beta| = \sum_{i=1}^D \beta_i$, and $w_\beta$ is the coefficient of the corresponding term that depends on the module parameters. Furthermore, the degree of the polynomial grows exponentially with the model depth.
\end{lemma}}
The proof of Lemma 1 is in the Appendix. Notably, the output polynomial features contain high-order interactions of the input attribute, resulting in an architecture that combines explicit feature engineering of attributes with implicit functions in deep neural networks.
\vspace{-4mm}
\subsubsection{Self-Gating} The second type of self-interaction we propose is self-gating. For self-gating, we set $g_a$ and $g_b$ as the ReLU and $g_s$ as the sigmoid activation:
\vspace{-0.5em}
\begin{align}
\small
\label{eq:self-gating}
\mathbf{a}_{\ell+1} = \mathrm{ReLU}\left(\Phi_a\left(\mathbf{a}_{\ell}\right)\right) \, *& \, \sigma\left(\Phi_s\left(\mathbf{a}_{\ell}\right)\right) \nonumber \\ &+ \mathrm{ReLU}\left(\Phi_b\left(\mathbf{a}_{\ell}\right)\right)
\end{align}

where $\sigma$ is the sigmoid activation function. The function $\Phi_s$ with a sigmoid activation maps the projected attribute to the range $[0,1]$, acting as a gate for each dimension of the function $\Phi_a$. Values closer to one signify the higher importance of a particular attribute dimension, while values closer to zero imply the opposite. Finally, function $\Phi_b$ projects the attribute to the same space as $\Phi_s$ and $\Phi_a$, resulting in a bias vector similar to the residual connection. We empirically find that $\Phi_a$ and $\Phi_s$ help learn a robust and global attribute vector while $\Phi_b$ helps stabilize model training. In our experiments, we found self-gating to perform better than polynomial-based self-interactions. Unless stated otherwise, we use self-gating modules in MAIN for the main results in experiments. In section~\ref{sec:pk-vs-sg}, we include an ablation comparing the two types of self-interactions.

\subsection{Inverse Regularization (IR)}
In the previous section, we proposed two types of SIA modules that map the semantic attribute space to the visual feature space. As noted in \ref{sec:background-notation}, the SIA-based attribute encoder is learned from only the seen class data. Because of the propensity for model bias towards the seen class samples, the unseen class attributes can be projected to the seen class features. To overcome this problem, we propose using inverse regularization that learns to project the resulting visual features from the attribute encoder back to the semantic space. We first provide a theoretical motivation for IR, which shows that IR is closely connected to entropy regularization in the visual feature space.
{\begin{lemma}[Maximize Entropy with 
IR] Let $t_\xi(\mathbf{a}|z) = \mathcal{N}(\mathbf{a};\mathcal{R}_\xi(z), I)$ be the probabilistic inverse map associated with the attribute encoder $f_\Phi$, where $z = f_\Phi(\mathbf{a})$ denotes the attribute embedding and $\mathcal{R}_\xi$ is the inverse regressor network with parameters $\xi$ that projects the attribute embedding in the visual feature space back to the semantic attribute space. Then, the mutual information between the attribute $\mathbf{a}$ and the attribute embedding $z$ is defined as:
\begin{equation}\small
I(\mathbf{a}; z) = H\left(z;\Phi\right) \geq H\left(\mathbf{a}\right) + \mathbb{E}_{\mathbf{a} \sim p(\mathbf{a})} \left[\log{t_\xi(\mathbf{a}|f_\Phi(\mathbf{a}))}\right].    
\end{equation}
\end{lemma}}
The proof is included in the Appendix. Since $H\left(a\right)$ is a constant with respect to ($\Phi$, $\xi$), maximizing $\mathbb{E}_{\mathbf{a} \sim p(\mathbf{a})} \left[\log{t_\xi(\mathbf{a}|f_\Phi(\mathbf{a}))}\right]$ increases the entropy $H\left(z;\Phi\right)$ in the embedding space. Hence, adding IR maximizes the entropy in the induced visual feature space, which acts as a regularization when training the attribute encoder and avoids over-fitting the model to seen classes.

The IR can be easily achieved by minimizing the cyclic-consistency loss between the semantic and visual space. The inverse regularization loss corresponding to Lemma 2 is defined as:
\begin{equation}
\small
\label{eq:cyclic_consistency}
\mathcal{L}_c = \sum_{\mathbf{a} \in \cA_{\leq t}^s} ||\mathcal{R}_\xi(f_\Phi(\mathbf{a})) - \mathbf{a}||^2
\end{equation}

Few recent works~\cite{vermageneralized,felix2018multi,kodirov2015unsupervisedDA} have also explored similar regularization for improved performance. However, a theoretical justification of how IR for the attribute encoder is connected to entropy maximization in the visual feature space is one of the contributions of this work. In our experiments, we show that adding entropy regularization improves the model's generalization to unseen classes.\\
\noindent \textbf{Overall training loss:} For the classification task, the probabilities are computed as $p(t=y|x) = e^{s_y}/\sum_{y'}{e^{s_{y'}}}$, where $s_y = {f_\Phi(\mathbf{a}_y)}^T x$ is the class logit and $t$ is predicted label. We use the standard cross entropy ($H$) as the loss:\vspace{-2mm}
\begin{equation}
\label{eq:cross_entropy}
\mathcal{L}_s=H\left(p\left(t=y|x\right), e_y\right)
\end{equation}
where $e_y$ is the one-hot representation for the ground-truth label $y$ corresponding to the image embedding $x$ in the visual feature space. We optimize the joint loss obtained from Equations~\ref{eq:cyclic_consistency} and \ref{eq:cross_entropy}:\vspace{-3mm}
\begin{equation}
\label{eq:final_main_loss}
\mathcal{L}_{\tau}=\mathcal{L}_s+\lambda\mathcal{L}_c
\end{equation}
where $\tau$ is the sampled batch and $\lambda$ is a hyperparameter tuned on the validation data. 
  
\vspace{-3mm}
\subsection{Reservoir Sampling} 
\vspace{-2mm}
In continual learning settings, we assume that tasks arrive sequentially, such that at the $i^{th}$ step, training samples are only available from the task $T_i$, and samples from previous tasks ($T_1,\dots, T_{t-1}$) are not accessible. Due to the tendency of neural networks to experience catastrophic forgetting~\cite{mccloskey1989catastrophic}, the model is likely to forget previously learned knowledge when learning new tasks. To mitigate catastrophic forgetting, MAIN incorporates reservoir sampling~\cite{vitter1985random} using a small and constant memory size to store selected samples from previous tasks and replaying these samples while training the attribute encoder on the current task. This has several advantages: ($i$) each task can be trained with constant time and computational resources, and ($ii$) the size of the memory does not grow as the tasks increase. Replaying samples from the reservoir effectively mitigates forgetting. We select each sample for replay with probability $M/N$, where $M$ is the memory budget and $N$ is the number of samples selected so far.
\vspace{-0.5em}

\subsection{Training with Meta-Learning}
\label{sec:meta}
While reservoir sampling is effective for mitigating forgetting, it also has several drawbacks. As the task number grows, a constant memory budget means the number of samples of each class diminishes, as the same amount of memory has to accommodate a large number of classes. Similarly, the current task will have a sizably larger number of samples for each class from the current task than the number of samples in the reservoir for classes of past tasks. Therefore, training can be difficult, resulting in models that generalize unevenly across all tasks/classes. 
This issue is similar to the problem of few-shot learning, where we have to learn a model using only a few training examples present in the reservoir from the previous tasks. Meta-learning models have shown promising results in the few-shot learning settings~\cite{finn2017model,nichol2018first}. Recent work \cite{vinyals2016matching,snell2017prototypical,raghu2019rapid} shows that even without any fine-tuning meta-learned models generalize well to novel classes.

Inspired by the recent success of meta-learning, MAIN leverages Reptile~\cite{nichol2018first}, a first-order meta-learning approach to train the attribute encoder. The training does not require storing gradients in memory for the inner loop, resulting in fast training of a generalized model. Given an attribute encoder $f_\Phi$ parameterized with $\Phi$, we define $U_\tau^m$ as the operator representing $m$ number of gradient updates of the loss function (\ref{eq:final_main_loss}) with respect to $\Phi$. With Reptile, the training batch $\tau$ is sampled from an augmented dataset containing the training samples from the current task and the stored memory. The encoder parameters are updated as follows:
\begin{equation}
\label{eq:meta_finalupdate}
\Phi\leftarrow\Phi-\eta(\Phi-\widetilde{\Phi}),
\end{equation}
where $\widetilde{\Phi} = U_\tau^m(\Phi)$. The algorithm~\ref{alg:reptile_mczsl} shows the Reptile-based training for continual ZSL with an augmented dataset. As we later show in the ablations, meta-learning significantly improves various experimental settings considered for both continual and generalized ZSL. To the best of our knowledge, this is the first work to show the merits of meta-learning in the continual ZSL setting.
\begin{algorithm}
\caption{Reptile~\cite{nichol2018first} for continual ZSL on a task $t$.}\label{alg:reptile_mczsl}
\begin{algorithmic}[1]
    \State Given encoder $f_\Phi$, task $\mathcal{D}_t$, memory $\mathcal{C}$, \& $k>1$.
    \For{iteration $= 1, 2, \dots$}                    
    \State {Sample task $\tau$ from the augmented dataset $\mathcal{C} \cup \mathcal{D}_t$}.
    \State Compute $\Tilde{\Phi} = U_\tau^k(\Phi)$, denoting $k$ steps of SGD starting at $\Phi$.
    \State Update: $\Phi \gets \Phi + \epsilon (\Tilde{\Phi} - \Phi)$.
    \EndFor
\end{algorithmic}
\end{algorithm}
\vspace{-1.0em}

\begin{table*}[t]
\scriptsize
\vspace{-3mm}
    \centering
    \caption{Mean seen accuracy (mSA), mean unseen accuracy (mUA), and their harmonic mean (mH) for fixed continual GZSL.}
    \vspace{-4mm}
     \addtolength{\tabcolsep}{0.4pt}
    \label{tab:gen_res_S1}
        \begin{tabular}{l|ccc|ccc|ccc|ccc|ccc}
            \toprule
            & \multicolumn{3}{c|}{CUB} & \multicolumn{3}{c|}{aPY} & \multicolumn{3}{c|}{AWA1} & \multicolumn{3}{c|}{AWA2} & \multicolumn{3}{c}{SUN} \\
            \cline{2-16}          &
            \multicolumn{1}{c}{mSA} & \multicolumn{1}{c}{mUA} & \multicolumn{1}{c|}{mH} & \multicolumn{1}{c}{mSA} & \multicolumn{1}{c}{mUA} & \multicolumn{1}{c|}{mH} & \multicolumn{1}{c}{mSA} & \multicolumn{1}{c}{mUA} & \multicolumn{1}{c|}{mH} & \multicolumn{1}{c}{mSA} & \multicolumn{1}{c}{mUA} & \multicolumn{1}{c|}{mH} & \multicolumn{1}{c}{mSA} & \multicolumn{1}{c}{mUA} & \multicolumn{1}{c}{mH} \\
            \midrule
            Seq-CVAE~\cite{mishra2017generative} & 24.66 & 8.57  & 12.18 & 51.57 & 11.38 & 18.33 & 59.27 & 18.24 & 27.14 & 61.42 & 19.34 & 28.67 & 16.88 & 11.40 & 13.38 \\	Seq-CADA~\cite{schonfeld2019generalized} & 40.82 & 14.37 & 21.14 & 45.25 & 10.59 & 16.42 & 51.57 & 18.02 & 27.59 & 52.30 & 20.30 & 30.38 & 25.94 & 16.22 & 20.10 \\
            CZSL-CA+res~\cite{gautam2020generalized} & 43.96 & 32.77 & 36.06 & 57.69 & 20.83 & 28.84 & 62.64 & 38.41 & 45.38 & 62.80 & 39.23 & 46.22 & 27.11 & 21.72 & 22.92 \\
            NM-ZSL~\cite{skorokhodov2021class} & 55.45 & 43.25 & 47.04 & 45.26 & 21.35 & 27.18 & 70.90 & 37.46 & 48.75 & 76.33 & 39.79 & 51.51 & 50.01 & 19.77 & 28.04 \\
            GRCZSL~\cite{gautam2021generative}& 41.91& 14.12& 20.48& 66.47& 12.06& 20.12& 78.66& 21.86& 33.56& 81.01& 23.59& 35.82& 17.74& 11.50& 13.73\\
            CARNet~\cite{gautam2022refinement}& 43.41 &47.44 &44.23& ----& ----& ----& ----& ----& ----& ----& ----& ----& 23.64 &30.95 &26.03\\
            ULTC~\cite{kuchibhotla2022unseen}&42.39& 37.09 &36.35&   60.96 &26.50 &33.96 &66.30 &50.63 &\textbf{55.59}& 67.15& 54.19 &\textbf{61.32}&  21.44& 27.57& 23.87\\
            \midrule
            MAIN w/o IR & 52.27 &48.59& 49.02 & 52.02&26.04&34.30 & 65.86 & 45.66 &52.33& 64.43 & 48.20 & 54.45  & 48.53& 24.31& 31.86\\
            MAIN (Ours) & 58.58 & 47.81  & \textbf{51.43} &62.81& 25.17 & \textbf{35.42} & 65.15 & 47.05 &{53.34}& 64.02 & 49.60 & {55.36}  & 48.50& 25.27& \textbf{32.72}  \\
            \bottomrule
        \end{tabular}%
\end{table*}
\begin{table*}[t]
\scriptsize
\vspace{-3mm}
    \centering
    \caption{Mean seen accuracy (mSA), mean unseen accuracy (mUA), and  harmonic mean (mH) for dynamic continual GZSL.}
    \vspace{-4mm}
\addtolength{\tabcolsep}{0.4pt}
        \begin{tabular}{l|ccc|ccc|ccc|ccc|ccc}
            \toprule
            & \multicolumn{3}{c|}{CUB} & \multicolumn{3}{c|}{aPY} & \multicolumn{3}{c|}{AWA1} & \multicolumn{3}{c|}{AWA2} & \multicolumn{3}{c}{SUN}\\
            \cline{2-16}          & \multicolumn{1}{c}{mSA} & \multicolumn{1}{c}{mUA} & \multicolumn{1}{c|}{mH} & \multicolumn{1}{c}{mSA} & \multicolumn{1}{c}{mUA} & \multicolumn{1}{c|}{mH} & \multicolumn{1}{c}{mSA} & \multicolumn{1}{c}{mUA} & \multicolumn{1}{c|}{mH} & \multicolumn{1}{c}{mSA} & \multicolumn{1}{c}{mUA} & \multicolumn{1}{c|}{mH} & \multicolumn{1}{c}{mSA} & \multicolumn{1}{c}{mUA} & \multicolumn{1}{c}{mH} \\
            \midrule
            Seq-CVAE~\cite{mishra2017generative} & 38.95 & 20.89 & 26.74 & 65.87 & 17.90 & 25.84 & 70.24 & 28.36 & 39.32 & 73.71 & 26.22 & 36.30 & 29.06 & 21.33 & 24.33 \\
            Seq-CADA~\cite{CADA-VAE} & 55.55 & 26.96 & 35.62 & 61.17 & 21.13 & 26.37 & 78.12 & 35.93 & 47.06 & 79.89 & 36.64 & 47.99 & 42.21 & 23.47 & 29.60 \\
            CZSL-CV+res~\cite{gautam2020generalized} & 63.16 & 27.50 & 37.84 & 78.15 & 28.10 & 40.21 & 85.01 & 37.49 & 51.60 & 88.36 & 33.24 & 47.89 & 37.50 & 24.01 & 29.15\\
            CZSL-CA+res~\cite{gautam2020generalized} & 68.18 & 42.44 & 50.68 & 66.30 & 36.59 & 45.08 & 81.86 & 61.39 & 69.92 & 82.19 & 55.98 & 65.95 & 47.18 & 30.30 & 34.88 \\
            NM-ZSL~\cite{skorokhodov2021class} & 64.91 & 46.05 & 53.79 & 79.60 & 22.29 & 32.61 & 75.59 & 60.87 & 67.44 & 89.22 & 51.38 & 63.41 & 50.56 & 35.55 & 41.65 \\
            GRCZSL~\cite{gautam2021generative}& 59.27& 26.03& 35.67& 77.61& 22.26& 33.01& 87.88& 30.34& 44.32& 90.45& 36.66& 51.74& 30.78& 22.59& 25.54\\
            UCLT~\cite{kuchibhotla2022unseen}&50.51& 52.2& 51.18&  74.92 &33.94& \textbf{46.26}& 79.51& 69.13 &73.49&79.19 &70.71 &\textbf{74.09}& 45.43& 44.59& 44.56\\
            \hline
            MAIN w/o IR  & 61.73 &71.14  & 65.94 & 55.41 & 39.29 & 43.51 & 82.37& 64.99 &{72.11}& 86.79 &60.82& {70.52} & 46.26 & 50.70 &{48.25}\\
            MAIN (Ours)  & 64.59 &70.40  & \textbf{67.20} & 57.50 & 38.59 & {45.10} & 80.56& 69.05 &\textbf{73.83}& 85.56 &65.59& {73.45} & 47.25 & 52.01 &\textbf{49.39}\\
            \bottomrule
    \end{tabular}
    \label{tab:gen_res_S2}
    \vspace{-5mm}
\end{table*}
\section{Experiments}
While the primary focus of this work is on continual ZSL settings ($i.e.$, Fixed continual~\cite{skorokhodov2021class} and Dynamic continual~\cite{gautam2020generalized}), we also analyze and compare MAIN with several baselines on the non-continual and standard ZSL and GZSL settings. In supplementary, we discuss these different settings and the corresponding evaluation protocols in detail. In section~\ref{sec:results}, we show that MAIN outperforms recent continual GZSL baselines while also achieving state-of-the-art results on the standard GZSL and ZSL settings. Furthermore, the training of MAIN is $\sim100\times$ faster than other alternatives which require training generative models. Additional implementation details are provided in the supplementary.

\subsection{Datasets and Baselines}
\vspace{-2mm}
\label{sec:dataset}
We conduct experiments on five widely used datasets for ZSL. In our experiment we used CUB-200~\cite{CUB}, AWA2~\cite{xian2018zero},  AWA1~\cite{AWA1}, SUN~\cite{patterson2012sun} and aPY~\cite{aPY} datasets. Further details about the datasets are in the supplementary.

\vspace{-0.7em}
\subsubsection{Fixed and Dynamic Continual GZSL Tasks}
\begin{figure}
\vspace{-3mm}
    \centering
    \includegraphics[scale=0.265]{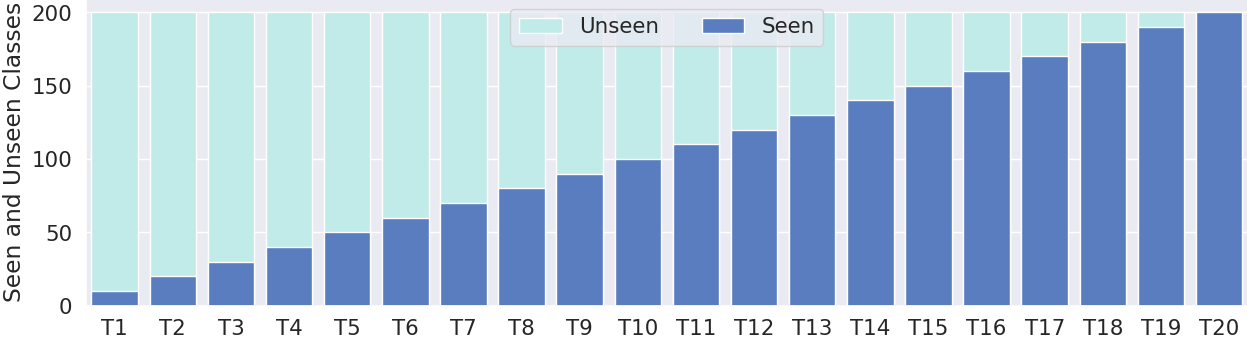}
    \includegraphics[scale=0.265]{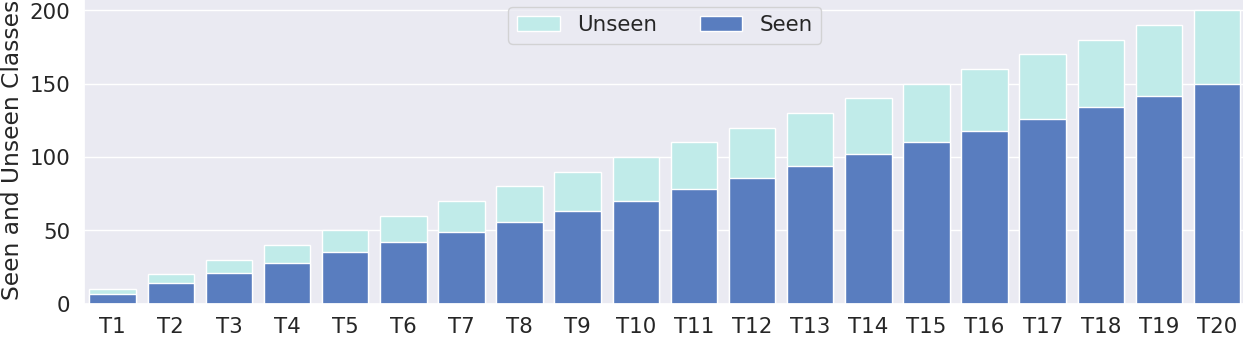}
    \vspace{-5mm}
    \caption{Seen and Unseen class distribution for the \emph{fixed} and \emph{dynamic} GZSL setting for the CUB dataset. In the fixed model as task proceed unseen classes decrease, however in dynamic GZSL as the task increase seen unseen increase. In both with the increase of the task forgetting increases.}
    \label{fig:fixed-dynamic}
    \vspace{-5mm}
\end{figure}

\vspace{-0.7em}

In the fixed continual generalized zero-shot learning (GZSL) setting, we assume that we have a fixed number of classes that arrive sequentially. Figure-\ref{fig:fixed-dynamic} (top) illustrates the seen-unseen distribution for the CUB dataset, where the data is divided into 20 tasks. Initially, the problem is highly challenging as only a few classes are visible, while the remaining classes are unseen. As the tasks progress, the number of seen classes grows, and the number of unseen classes decreases, as the latter are added to the seen classes.

The dynamic continual GZSL setting is more practical, where both seen and unseen classes increase as the tasks advance. Figure-\ref{fig:fixed-dynamic} (bottom) depicts the dynamic setting for the CUB dataset. Here, 150 seen classes and 50 unseen classes are divided into 20 tasks, with $[7,7,7,7,7,7,7,7,7,7,8,8,8,8,8,8,8,8,8,8]$ and $[3,3,3,3,3,3,3,3,3,3,2,2,2,2,2,2,2,2,2,2]$ classes in each task for seen and unseen data, respectively. In this scenario, both seen and unseen classes grow with each task, but only the data from the current task is available for training. For instance, in the $3^{rd}$ task, there are $21$ seen classes ($7+7+7$) and $9$ unseen classes ($3+3+3$), yet only the data from the $3^{rd}$ task's seven classes are accessible for training. The reservoir memory for each dataset is set to $B \times \#total\_classes$, where $B$ represents the reservoir parameter, indicating the number of images per class. Specifically, the reservoir sample $B$ is set to $25$ for AWA1, AWA2, and aPY datasets, $10$ for CUB dataset, and $5$ for the SUN dataset. Further details for other datasets can be found in the supplementary materials.

\vspace{-1em}
\subsubsection{Continual ZSL Baselines}
This section describes the baselines used for the continual ZSL setting. For the non-continual (standard) ZSL and GZSL baselines, please refer directly to tables~\ref{tab:gzsl} and \ref{tab:ZSL} for a variety of non-continual baselines we considered. To the best of our knowledge, continual ZSL has mainly been explored in \cite{gautam2020generalized, skorokhodov2021class, wei2020lifelong, chaudhry2018efficient,kuchibhotla2022unseen}. For comparison, we consider CZSL-CV+res and CZSL-CA+res~\cite{gautam2020generalized}, which use conditional variational auto-encoders (VAE) and CADA~\cite{schonfeld2019generalized} with a memory reservoir, respectively. We also consider the recently proposed GRCZSL~\cite{gautam2021generative} that uses generative replay to overcome catastrophic forgetting in the continual GZSL setting. NM-ZSL~\cite{skorokhodov2021class} is also a competitive continual ZSL baseline and other continual ZSL baselines include Seq-CVAE~\cite{mishra2017generative} and Seq-CADA~\cite{schonfeld2019generalized}, which
sequentially train a VAE over tasks to augment the training dataset using replay. The recent work UCLT~\cite{kuchibhotla2022unseen} uses a generative model and reply to solve the CZSL problem.

\begin{table*}[ht]
\scriptsize
    \centering
    \caption{Mean seen accuracy (mSA), mean unseen accuracy (mUA), and their harmonic mean (mH) for GZSL. \textbf{Top:} Methods similar to the proposed non-generative model where the model does not require the unseen class attributes. \textbf{Bottom:} Expensive generative model approaches, which require unseen class attributes during training. w/o IR: no inverse regularization.}
    \vspace{-2mm}
    \label{tab:gzsl}%
    \addtolength{\tabcolsep}{2.2pt}
        \begin{tabular}{l|ccc|ccc|ccc|ccc|c}
            \toprule
            & \multicolumn{3}{c|}{SUN} & \multicolumn{3}{c|}{CUB} & \multicolumn{3}{c|}{AWA1} & \multicolumn{3}{c|}{AWA2} & \multicolumn{1}{c}{Train Time}
            \\
            \cline{2-13}
            & \multicolumn{1}{c}{mSA} & \multicolumn{1}{c}{mUA} & \multicolumn{1}{c|}{mH} & \multicolumn{1}{c}{mSA} & \multicolumn{1}{c}{mUA} & \multicolumn{1}{c|}{mH} & \multicolumn{1}{c}{mSA} & \multicolumn{1}{c}{mUA} & \multicolumn{1}{c|}{mH} & \multicolumn{1}{c}{mSA} & \multicolumn{1}{c}{mUA} & \multicolumn{1}{c|}{mH} & 
            \\
            \midrule
            DEM ~\cite{dem} & 34.3& 20.5& 25.6& 57.9& 19.6& 29.2& 84.7& 32.8& 47.3& 86.4& 30.5& 45.1&20 minutes\\
            CRNet~\cite{zhang2019co}&36.5& 34.1& 35.3& 56.8& 45.5& 50.5& 78.8& 52.6& 63.1& 74.7& 58.1& 65.4&--\\
            LFGAA~\cite{liu2019attribute}&34.9& 20.8& 26.1&79.6& 43.4& 56.2&--&--&--&90.3& 50.0& 64.4&--\\
            NM-ZSL~\cite{skorokhodov2021class} & 44.7 &41.6& 43.1& 49.9& 50.7& 50.3& 63.1& 73.4& 67.8& 60.2& 77.1& 67.6& 30 seconds  \\
            \midrule
            CVC-ZSL~\cite{li2019rethinking} & 36.3 &42.8& 39.3 &47.4& 47.6& 47.5& 62.7& 77.0& 69.1& 56.4& 81.4& 66.7& 3 hours \\
            SGAL~\cite{yu2019zero} & 42.9& 31.2& 36.1& 47.1& 44.7& 45.9& 52.7& 75.7& 62.2& 55.1& 81.2& 65.6& 50 minutes  \\
            TF-VAEGAN~\cite{narayan2020latent} & 45.6& 40.7& 43.0& 52.8& 64.7& 58.1& --& --& --& 59.8& 75.1& 66.6& 1.75 Hours \\
            LsrGAN~\cite{vyas2020leveraging} & 44.8& 37.7& 40.9& 48.1& 59.1& 53.0& --& --& --& 54.6& 74.6& 63.0& 1.25 hours  \\
            F-VAEGAN-D2~\cite{vaegan} & 45.1 &38.0& 41.3& 48.4& 60.1& 53.6& --& --& --& 57.6& 70.6& 63.5& --  \\
            ZSML~\cite{verma2019meta} & 45.1&21.7& 29.3& 60.0& 52.1& 55.7& 57.4& 71.1& 63.5& 58.9& 74.6& 65.8&3 hours  \\
            DAZLE~\cite{huynh2020fine}&24.3 &52.3 &33.2& 59.6& 56.7& 58.1&--&--&--&75.7 &60.3& 67.1\\
            CARNet~\cite{gautam2022refinement}&49.4& 40.5& 44.5&65.0& 59.6 &61.2&69.5& 74.7& 72.0 & 65.7& 79.7 &72.0\\
            TDCSS~\cite{feng2022non} &--&--&--& 44.2& 62.8 &51.9& 54.4 &69.8 &60.9&59.2& 74.9 &66.1&\\
            TransZero~\cite{chen2022transzero} & 52.6 &33.4& 40.8& 69.3 &68.3& \textbf{68.8}& 60.3 &81.1 &69.2& 61.3 &82.3 &70.2&\\
            ICCE~\cite{kong2022compactness} &--&--&--&67.3 &65.5 &66.4&67.4& 81.2& 73.6&65.3 &82.3 &72.8&\\ 
            \midrule
            MAIN w/o IR & 40.3 &46.9& {43.4}& 57.2& 66.4& {61.4}& 78.9&64.6& \textbf{71.8}& 77.9& 67.1& \textbf{72.1}& \textbf{30 seconds}  \\
            MAIN (Ours) & 40.0 &51.1& \textbf{44.8}& 58.7& 65.9& {62.1}& 77.9&71.9& \textbf{74.8}& 81.8& 72.1& \textbf{76.7}& 45 seconds\\
            \bottomrule
        \end{tabular}%
    \vspace{-5mm}
\end{table*}%
\subsection{Results}
\label{sec:results}
\vspace{-3mm}
\subsubsection{Fixed Continual GZSL}
\begin{figure}[h]
	\centering
    \includegraphics[width=4.1cm]{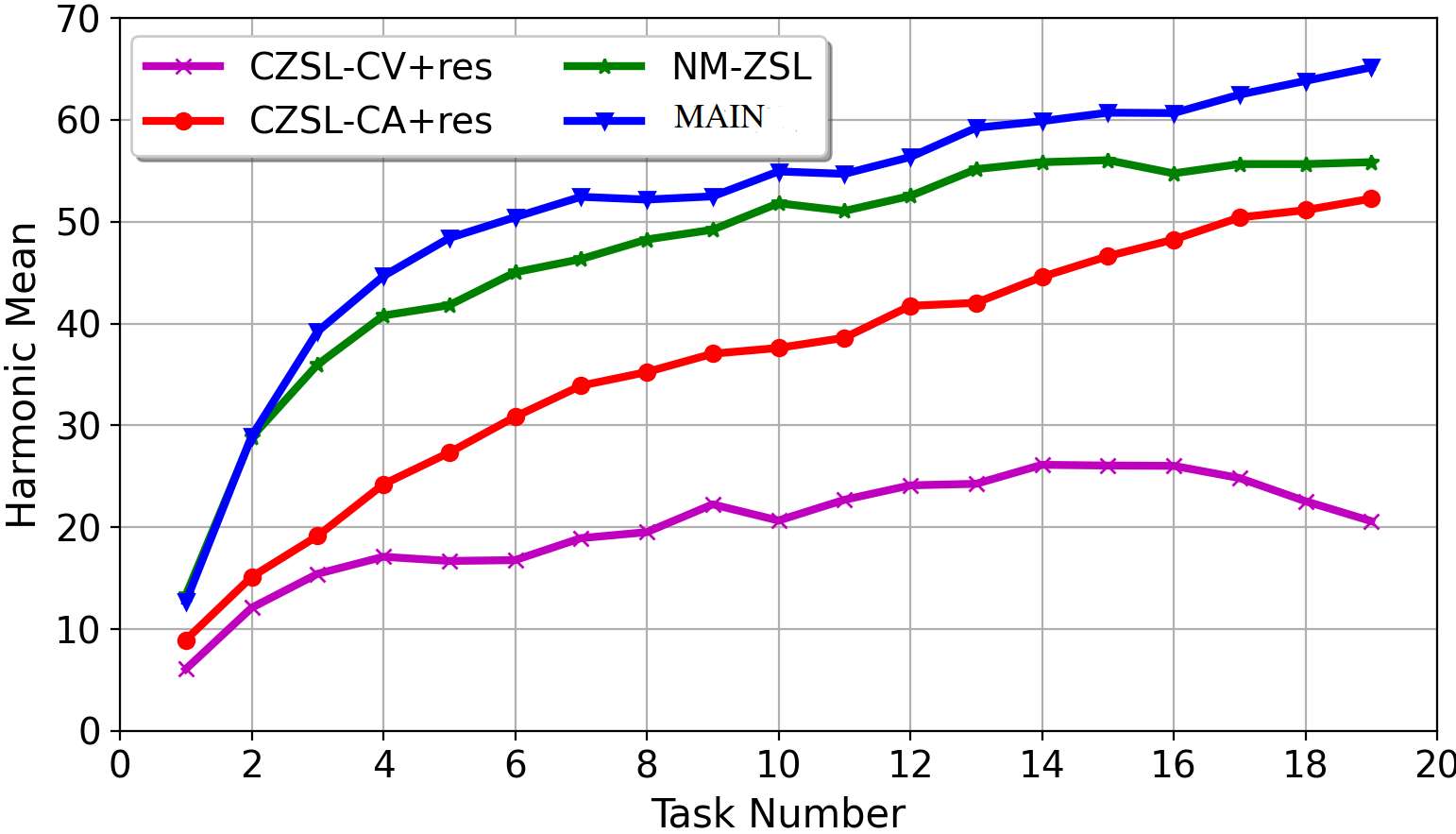}
	\includegraphics[width=4.1cm]{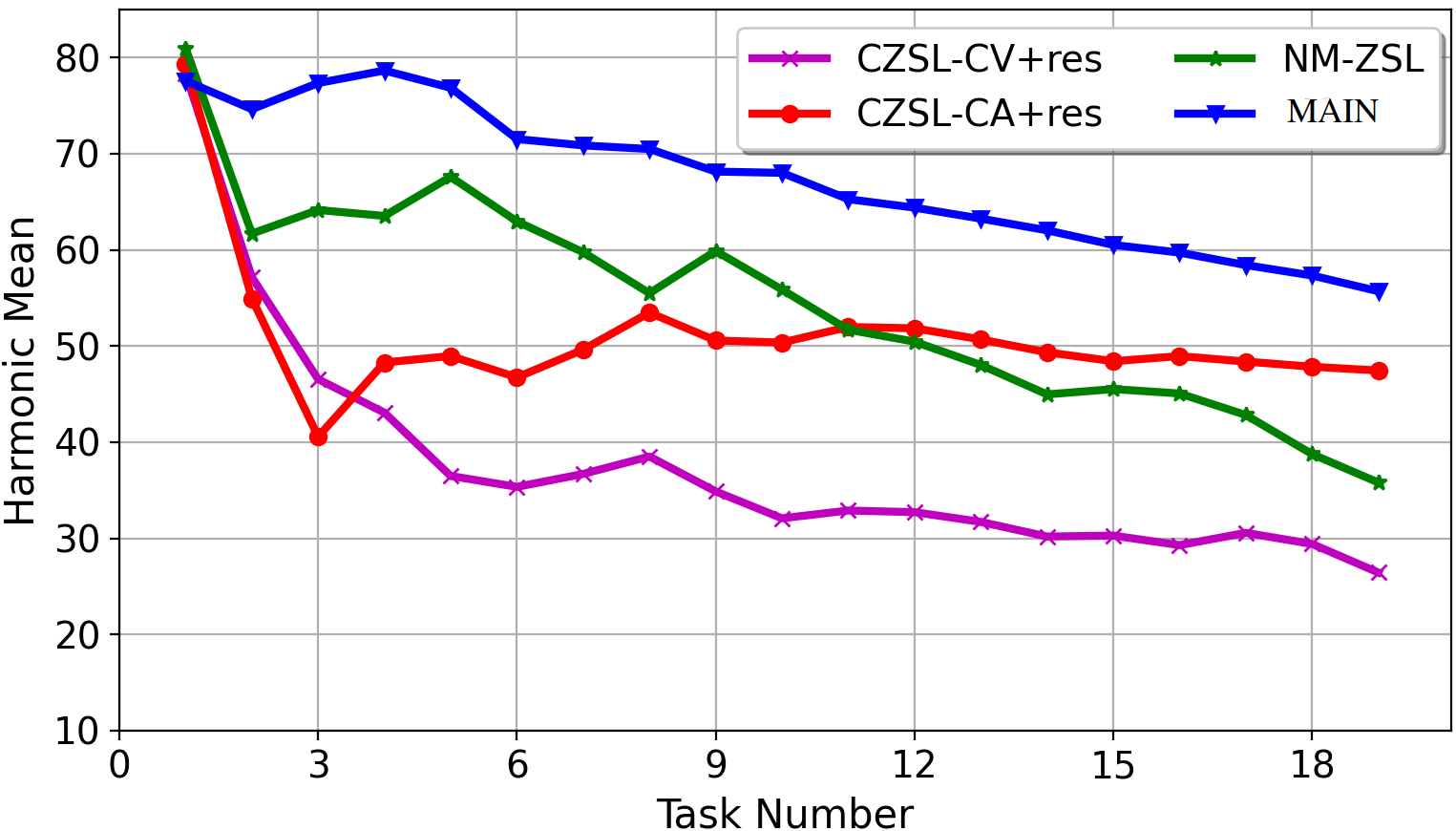}
	\caption{Harmonic mean per task for continual GZSL on the CUB dataset. \textbf{Left:} Represents the Fixed Continual GZSL setting. \textbf{Right:} Represents the Dynamic Continual GZSL setting.}
	\label{fig:hmean_pertask}
	\vspace{-1.0em}
\end{figure}

\vspace{-0.5em}
In the Fixed Continual GZSL~\cite{skorokhodov2021class} setting, as the model is trained on a sequence of tasks, the number of seen classes increases while the number of unseen classes decreases. The results for the CUB, aPY, AWA1, AWA2, and SUN datasets are shown in table~\ref{tab:gen_res_S1}. We use the same memory buffer size as in~ \cite{gautam2020generalized} for a fair comparison. On CUB, aPY, and SUN, the proposed model shows $4.39\%$, $1.46\%$, and $4.68\%$ absolute increase over the best baseline. However, compared to AWA1 and AWA2, we have a marginal difference. We also analyze the performance of various methods across each task in the CUB dataset in Figure~\ref{fig:hmean_pertask}. We observe that our proposed MAIN consistently outperforms recent baselines. Notably, these significant improvements are achieved without using any costly generative models.
\vspace{-4mm}
\subsubsection{Dynamic Continual GZSL} 
\vspace{-3mm}
An alternate framing of continual GZSL is explored in the Dynamic Continual GZSL~\cite{gautam2020generalized} setting, where each task consists of its own disjoint set of seen and unseen classes. In this setting, both the seen and unseen classes increase with $t$. As in fixed continual GZSL, we conduct experiments in the dynamic continual GZSL on CUB, aPY, AWA1, AWA2, and SUN, as shown in table~\ref{tab:gen_res_S2}. We observe that our proposed MAIN performs well, seeing absolute gains of $13.41\%$, $0.34\%$, and $4.83\%$ on the CUB, AWA1, and SUN datasets while being relatively competitive on aPY and AWA2. In figure~\ref{fig:hmean_pertask}, we show the harmonic mean per task for the CUB dataset. This is a harder setting compared to the fixed continual GZSL setting since each task brings more seen and unseen classes, making the classification task challenging due to more classes to distinguish from and increasing the opportunity for catastrophic forgetting of previous tasks; thus, accuracy tends to drop with more tasks. Regardless, we again observe that the proposed model consistently outperforms recent baselines.

\vspace{-5mm}
\subsubsection{Generalized Zero-Shot Learning}
\vspace{-3mm}
We conduct experiments on CUB-200, SUN, AWA1, and AWA2 datasets in the GZSL setting, reporting the accuracy over the seen and unseen classes in table~\ref{tab:gzsl}. We observe significant improvements using MAIN over previous methods on the SUN, AWA1, and AWA2 datasets, with absolute gains of  $0.3\%$, $1.2\%$, and $3.9\%$, respectively. Notably, the training\footnote{Time to train on the whole dataset with an Nvidia GTX 1080Ti, code is available in supplementary.} of MAIN relative to the baseline approaches can be $100$-$300\times$ faster without requiring the unseen class attributes \emph{a priori}. The generative baselines require learning complex generative models ($e.g.$, VAE or GAN) to synthesize realistic samples, which can be expensive. Moreover, generative baselines require an additional assumption about the knowledge of unseen class attributes \textit{a priori} during training. We also evaluate our model on the standard ZSL setting; the results are shown in table~\ref{tab:ZSL}. Once again. we observe that MAIN significantly outperforms both non-generative and generative baselines.

\begin{table}[t]
	\vspace{-0.2em}
	\scriptsize
	\centering
	\caption{Results on the standard ZSL setting. \textbf{Top:} Non-generative methods (like ours). \textbf{Bottom:} Expensive generative methods.}
	\vspace{-3mm}
	\label{tab:ZSL}
	\addtolength{\tabcolsep}{5.0pt}
	\begin{tabular}{l|c|c|c|c}
		\toprule
		&  AWA1&AWA2&CUB&SUN\\
		\midrule
		LATEM ~\cite{xian2016latent}& 55.1& 55.8 &49.3 &55.3\\
		ALE ~\cite{akata2013label}& 59.9& 62.5& 54.9& 58.1\\
		RelationNet~\cite{relation} &68.2& 64.2& 55.6& --\\
		SAE ~\cite{SAE}& 53.0& 54.1& 33.3& 40.3\\
		EF-ZSL~\cite{verma2017simple}&57.0&57.4&44.7&63.3\\
		DCN ~\cite{DCN} &65.2& --& 56.2& 61.8\\
		\midrule
		SE-ZSL~\cite{vermageneralized}&69.5&69.2&59.6&63.4\\
		f-CLSWGAN~\cite{xian2018feature} & 68.2& --& 57.3& 60.8\\
		LisGAN ~\cite{CVPR19invarient} &70.6&--& 58.8& 61.7\\
		ADA~\cite{khare2020generative}&--&70.4&70.9&63.3\\
		f-VAEGAN-D2~\cite{vaegan} &71.1& --& 61.0& 65.6\\
		LFGAA ~\cite{liu2019attribute}& --& 68.1& 67.6& 62.0\\
		SP-AEN~\cite{Chen_2018_CVPR}&--&58.5&55.4&59.2\\
		OCD-CVAE~\cite{keshari2020generalized}&--&71.3&60.3&63.5\\
        CARNet~\cite{gautam2022refinement}&75.0&73.7&73.1&63.1\\
        TransZero~\cite{chen2022transzero} &71.9&70.1&\textbf{76.8}& 65.6\\
            \midrule
		MAIN w/o IR&\textbf{76.9}&\textbf{76.5}&{71.4}&64.8\\
		MAIN (Ours)&\textbf{79.0}&\textbf{76.8}&{73.7}&\textbf{65.8}\\
		\bottomrule
	\end{tabular}
	\vspace{-4mm}
\end{table}

\vspace{-3mm}
\section{Ablation Studies}
\vspace{-2mm}
\label{sec:ablations}
We conduct extensive ablation studies on the proposed model's different components, observing that each of the proposed components play a critical role. We show the effects of different components on the AWA1 and CUB datasets in the fixed continual GZSL setting, with more ablation studies for dynamic continual GZSL in the supplementary material. 

\vspace{-0.2em}
\subsection{Effect of Self-Gating on the Attribute}
\begin{figure}
	\centering
	\includegraphics[width=0.23\textwidth]{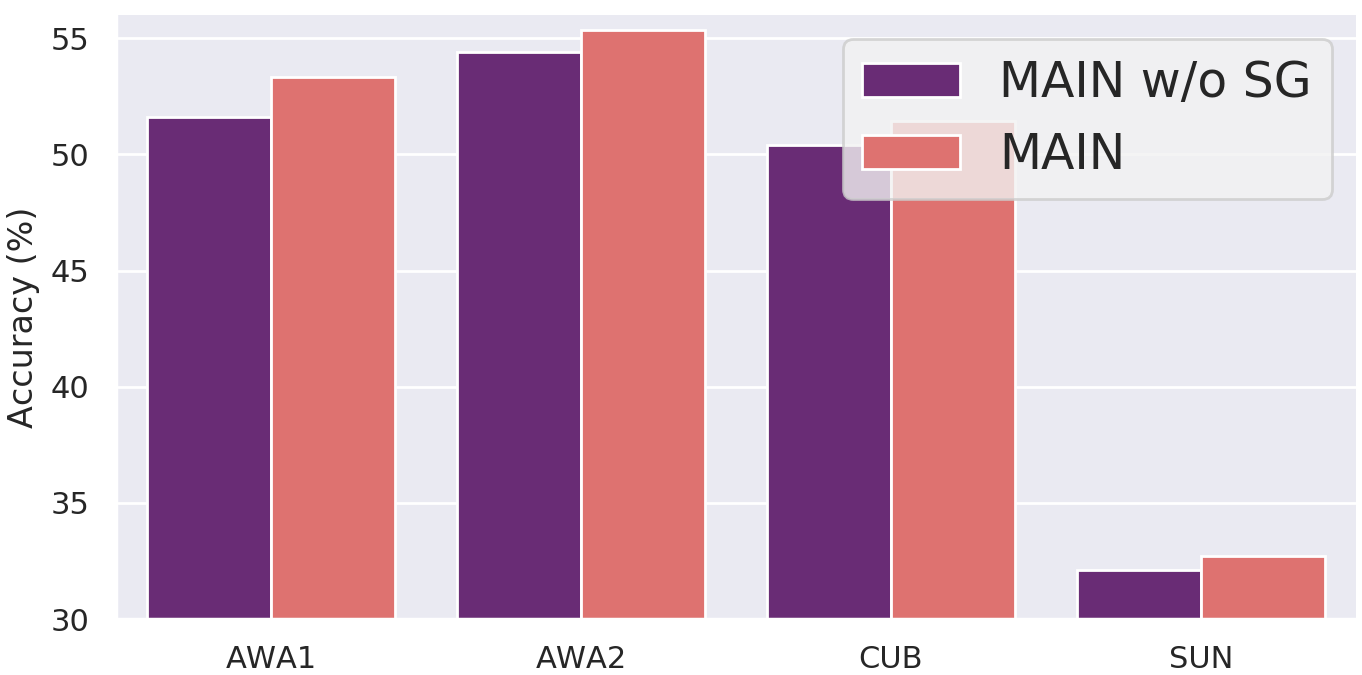}
	\includegraphics[width=0.23\textwidth]{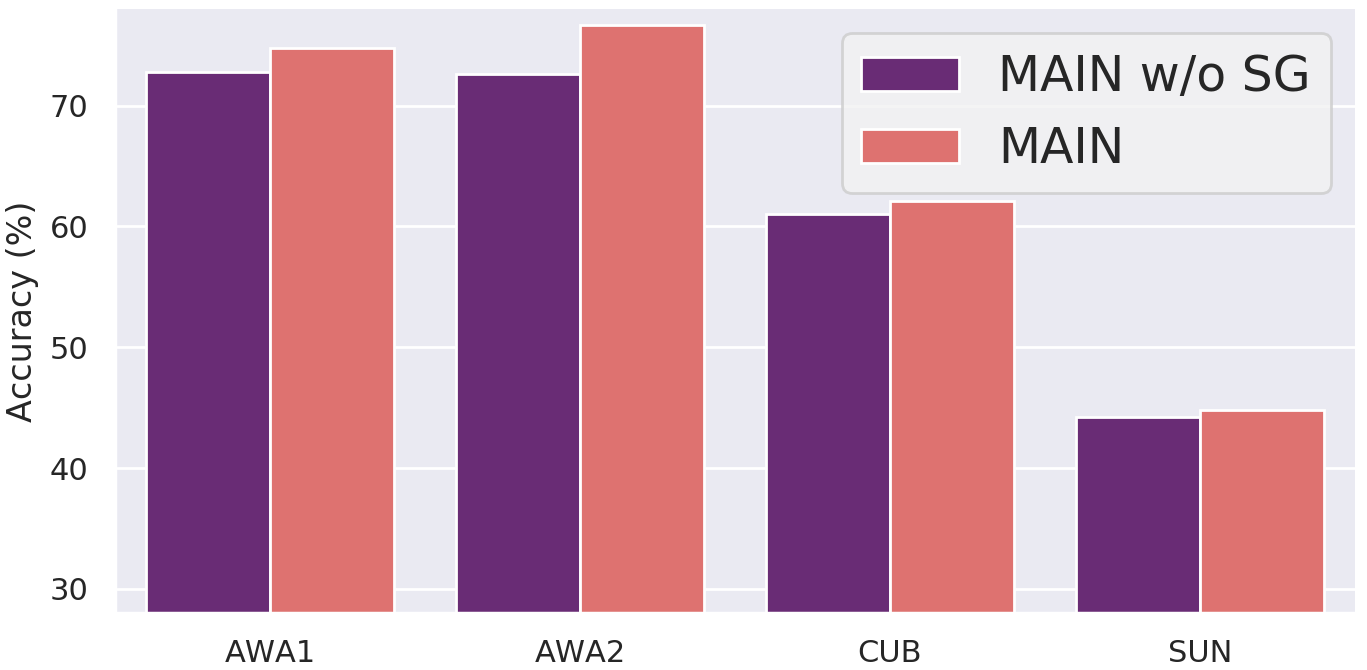}
	\vspace{-3mm}
	\caption{The proposed model with and without self-gating (SG) in the Fixed continual GZSL setting (\textbf{Left}) and the standard non-continual GZSL setting (\textbf{Right}).}
	\label{fig:self-gating}
	\vspace{-1.5em}
\end{figure}

\vspace{-0.2em}
\label{sec:self-gating}
As proposed in section~\ref{sec:self-interaction}, we use the self-gating module to learn the attribute embeddings. We perform an ablation to verify the importance of the self-gating module in figure~\ref{fig:self-gating}. As shown in the figure~\ref{fig:self-gating} (left), the performance (harmonic mean) decreases when we remove the self-gating on all the datasets. In particular, we observed that self-gating increased the harmonic mean by an average of $1.1$ on all the datasets. We observed similar gains in the standard GZSL setting (figure~\ref{fig:self-gating} (right)), where the self-gating led to an average improvement of $1.95$. This shows that self-gating is an important component of MAIN. 
\vspace{-3mm}
\subsection{Importance of Inverse Regularization (IR)}
\vspace{-2mm}
\label{sec:mmr}
The IR imposing cyclic consistency loss~(\ref{eq:cyclic_consistency}) plays a crucial role in the generalization capability of the attribute encoder to novel classes. In tables~\ref{tab:gzsl}, \ref{tab:gen_res_S1}, \ref{tab:gen_res_S2}, \ref{tab:ZSL}, we have shown the effect of IR. MAIN w/o IR shows the result without IR. We observe a consistent improvement from the cyclic consistency loss. Notably, even without IR, MAIN outperformed recent approaches by a significant margin.

\vspace{-2mm}
\subsection{Effect of Meta-training}
\vspace{-2mm}
\label{sec:meta-training}
\begin{figure}[!ht]
\vspace{-0.6em}
	\centering
	\includegraphics[width=0.23\textwidth]{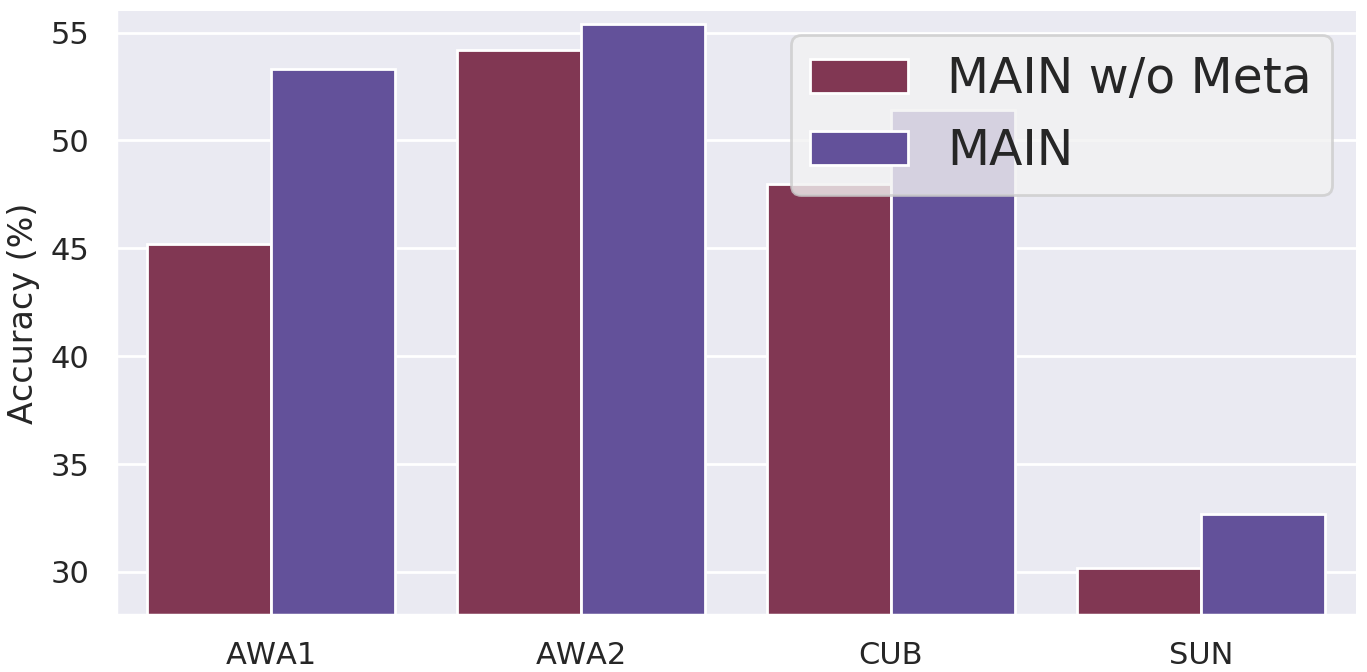}
	\includegraphics[width=0.23\textwidth]{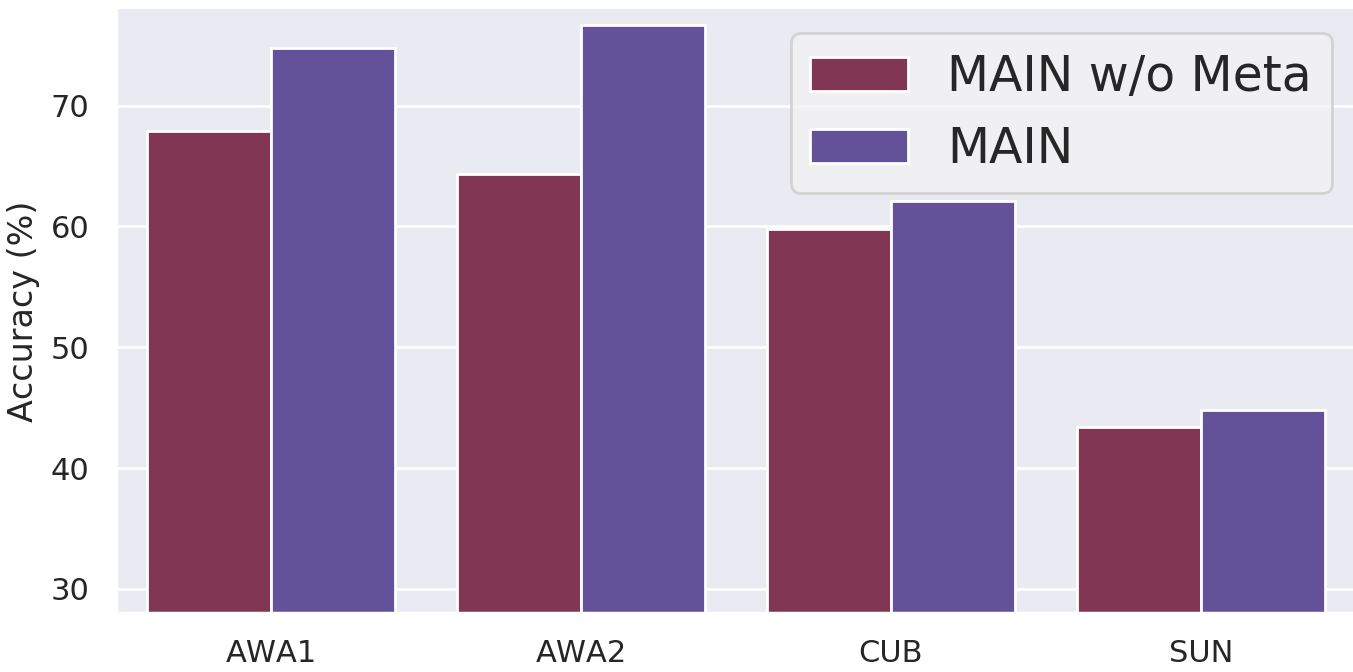}
	\vspace{-2mm}
	\caption{The proposed model with meta-learning and without meta-learning in the Fixed Continual GZSL setting (\textbf{Left}) and the standard non-continual GZSL setting (\textbf{Right}).}
	\vspace{-3mm}
	\label{fig:meta_ablation}
\end{figure}
While we used meta-learning in our proposed MAIN (See section~\ref{sec:meta}), the attribute encoder can also be learned directly by minimizing the loss in~(\ref{eq:final_main_loss}) using a standard optimizer without using meta-learning. We evaluate the model performance with and without meta-learning-based training on the fixed continual GZSL and the standard non-continual GZSL settings. The result is shown in figure~\ref{fig:meta_ablation}. We observed that if we withdraw the meta-learning-based training, the MAIN performance drops significantly. On AWA1, AWA2, CUB, and SUN datasets for fixed continual GZSL, the harmonic mean of MAIN drops from $53.3$ to $45.2$, $55.4$ to $54.2$, $51.4$ to $48.0$ and $32.7$ to $30.2$, respectively. Similarly, in the GZSL setting, MAIN achieves the harmonic mean of $74.8$, $76.7$, $62.1$, and $44.8$ for the AWA1, AWA2, CUB, and SUN datasets, respectively, which drops to $67.9$, $64.3$, $59.8$ and $43.4$ in the absence of meta-learning.
\vspace{-2mm}
\subsection{Polynomial Kernel vs. Self-gating}
\label{sec:pk-vs-sg}
\begin{figure}
\centering
\includegraphics[height=3cm,width=8cm]{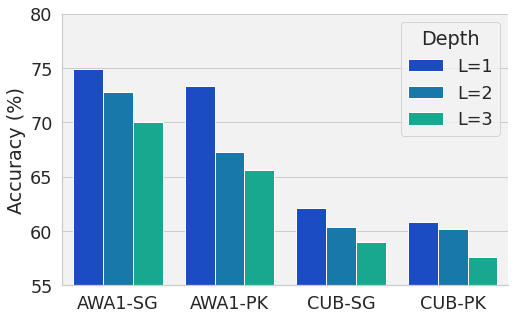}
\vspace{-3mm}
\caption{Comparison of the two types of Attribute self-Interaction modules as proposed in section~\ref{sec:self-interaction}; Polynomial Kernels (PK) vs. Self-Gating (SG). We also show the performance as a function of the depth of the self-interaction module with $L=\{1,2,3\}$. The only difference between SG and PK is that SG includes non-linear transformations.}
\vspace{-6mm}
\label{fig:sia}
\end{figure}

We performed the ablation over the two types of SIA modules proposed in~section \ref{sec:self-interaction} for varying depth $L=\{1,2,3\}$. The results are shown in figure-\ref{fig:sia}; here, we can observe that the self-gating module with $L=1$ has the best results. We found that using polynomial-based self-interaction leads to model overfitting due to an exponential rise in the degree with $L$. The only difference between the polynomial kernel and self-gating is, in the self-gating, we use non-linear transformation, and we observed which leads to better performance. Therefore if the polynomial kernel is equipped with the non-linearity, it can learn more complex functions with a smaller layer, and an increase in the layer may lead to overfitting because of the exponential increase in the complexity. In conclusion, for the $L=1$, the polynomial kernel with the non-linear projection is the best-performing model.
\vspace{-0.7em}

\vspace{-0.1em}
\section{Conclusions}
We propose the Meta-learned Attribute self-Interaction Network (MAIN), a method capable of operating in both generalized zero-shot learning and continual zero-shot learning settings.
Through first-order meta-learning-based training of the network and a novel self-interaction module for attributes, we obtain state-of-the-art results in GZSL settings across multiple common benchmarks. MAIN also incorporates a theoretically motivated regularization scheme for better generalization of the attribute encoder to the unseen class attributes. Notably, our approach does not rely on expensive generative models; this allows for considerably faster speed during training and relaxes the restrictive constraint requiring unseen class attributes to be known at training. Therefore, unlike methods relying on generative models, the inductive setting assumption in the attribute space makes the proposed model highly practical and easy to use.

Given clear connections between zero-shot and continual learning, we extend our approach to settings where class data arrive sequentially.
We adopt a data reservoir to mitigate catastrophic forgetting and combine it with meta-learning-based training, leading to a few-shot-based approach that naturally enables the model to efficiently learn from the few samples that can be saved in a buffer. Our results on the CUB, aPY, AWA1, AWA2, and SUN datasets in GZSL and two different protocols of continual GZSL demonstrate that MAIN outperforms a wide array of strong baselines. Ablation studies demonstrate that each component of our proposed approach is critical for its success.

{\small
\bibliographystyle{ieee_fullname}
\bibliography{egbib}
}

\appendix
\newpage
\newpage

\section*{Supplementary}

\section{Proofs}

{\begin{lemma}[Polynomial Approximation]
		Consider a model with $L$ layers of self-interaction modules with parameters $\{\Phi^\ell_a, \, \Phi^\ell_s, \, \Phi^\ell_b\}_{\ell=1}^L$ and identity activation $g^{\ell}_a(x) = g^{\ell}_s(x) = g^{\ell}_b(x) = x$. Let input to the model be:  $\mathbf{a} = [ a_{1}, a_{3}, \dots,  a_{D} ]$. Then, the output of the model $\mathbf{a}_L$ approximates following class of polynomial functions:
		\vspace{-0.7em}
		\begin{small}\begin{align}
				\label{eq:pol_approximation}
				\Bigg\{ P_{\ell}\left(\mathbf{a}\right) = \sum_\beta w_\beta \, a_{1}^{\beta_1} \, a_{2}^{\beta_2} \dots a_{D}^{\beta_D} \enspace \Bigg| \enspace 0 \leq |\beta| \leq 2^{\ell}
				\Bigg\},
		\end{align}\end{small}
		where the sum is across multiple terms (monomials), $\beta = [\beta_1,\dots,\beta_D]$ is a vector containing the exponents of each attribute in a given term having degree $|\beta| = \sum_{i=1}^D \beta_i$, and $w_\beta$ is the coefficient of the corresponding term that depends on the module parameters. Furthermore, the degree of the polynomial grows exponentially with the model depth.
\end{lemma}}
\noindent \textbf{Proof.} For the polynomial approximation, the self-interaction module is defined as:
\begin{align}
	\small
	\label{eq:self-interaction}
	\mathbf{a}_{\ell + 1} = \Phi_a^{\ell+1} \left(\mathbf{a}_\ell\right) * \Phi^{\ell+1}_s\left(\mathbf{a}_\ell\right) + \Phi^{\ell+1}_b\left(\mathbf{a}_\ell\right),
\end{align}
where $\mathbf{a}_0 = \mathbf{a} = [ a_{1}, a_{3}, \dots,  a_{D} ]$ and $*$ is the element-wise multiplication operator. For the analysis, we consider $\{\Phi^\ell_a, \, \Phi^\ell_s, \, \Phi^\ell_b\}_{\ell=1}^L \in \mathbb{R}^{D \times D}$ as square matrices for all $\ell \in [1, \dots, L]$, however, the analysis can be easily extended to the case when these matrices are rectangular. To show that the output of the $\ell^{th}$ self-interaction module approximates the polynomial, we show that each coordinate of $\mathbf{a}_\ell$ belongs to the class of polynomials (\ref{eq:pol_approximation}). In the following analysis, we denote the $i^{th}$ vector coordinate as $\mathbf{a}_\ell[i]$. Similarly, the $j^{th}$ column vector in the parameter matrix $\Phi^\ell$ is denoted as $\Phi^\ell[:, j]$. Then, the proof of the lemma follows from induction:\\

\noindent \textbf{Base:} Consider the base case for $\mathbf{a}_0 = [ a_{1}, a_{3}, \dots,  a_{D} ]$. Clearly, each coordinate $\mathbf{a}_0[i]$ belongs to the class of polynomials in ~(\ref{eq:pol_approximation}). In particular, $\mathbf{a}_0[i] = a_i $ $\in$ $\{P_{0}\left(\mathbf{a}\right)\}$.\\

\noindent \textbf{Induction step:} Assume that when $\ell = k$, $\mathbf{a}_k[i] \in \{P_k(a)\}$ $\forall i\in\{1,\dots,D\}$. Then, for $\ell = k+1$, we have:

\begin{small}
	\begin{align}
		\mathbf{a}_{k+1}& = \Phi^{k+1}_a (\mathbf{a}_k) * \Phi^{k+1}_s (\mathbf{a}_k) + \Phi^{k+1}_b (\mathbf{a}_k) \\
		&= \underbrace{\left(\sum_m \mathbf{a}_k[m] \, \Phi^{k+1}_a[:,m] \right)}_{r_a} * \underbrace{\left(\sum_n \mathbf{a}_k[n] \, \Phi^{k+1}_s[:,n]\right)}_{r_s} \nonumber \\ &\qquad \qquad + \underbrace{\left(\sum_m \mathbf{a}_k[m] \, \Phi^{k+1}_b[:,m]\right)}_{r_b} \label{eq:ri_notation}
	\end{align}
\end{small}
In~(\ref{eq:ri_notation}), each coordinate of the vectors $\{r_a, r_s, r_b\} \in \mathbb{R}^{D}$ belongs to the class of polynomials $\{P_k(a)\}$ since $\mathbf{a}_k[i] \in \{P_k(a)\}$ $\forall i\in\{1,\dots,D\}$. Then, the $i^{th}$ coordinate of $\mathbf{a}_{k+1}$ can be simplified as:
\begin{small}
	\begin{align}
		a_{k+1}[i] &= r_a[i] * r_s[i] + r_b[i], \text{ where}\\
		r_a[i] * r_s[i] &= \sum_m \sum_n \mathbf{a}_k[m] \, \mathbf{a}_k[n] \, \Phi^{k+1}_a[i,m] \, \Phi^{k+1}_s[i,n] \label{eq:exp_polynomial}\\
		\text{and } r_b[i] &= \sum_m \mathbf{a}_k[m] \, \Phi^{k+1}_b[i,m]
	\end{align}
\end{small}

Since~(\ref{eq:exp_polynomial}) is the sum of the product of polynomials, it follows that the resulting $\mathbf{a}_{k+1}[i]$ is a polynomial. Moreover, the degree of $\mathbf{a}_{k+1}[i]$ satisfies the following:
\begin{align}
	\text{deg}(\mathbf{a}_{k+1}[i]) &\leq \max_{m,n} \left[ \text{deg}(\mathbf{a}_{k}[m]) + \text{deg}(\mathbf{a}_{k}[n]) \right]\\
	&\leq 2^k + 2^k = 2^{k+1}
\end{align}
Hence, $\mathbf{a}_{k+1}[i]$ $\in$ $\{P_{k+1}(a)\}$ $\forall i\in\{1,\dots,D\}$.

{\begin{lemma}[Maximize Entropy with 
		IR]
		Let $t_\xi(a|z) = \mathcal{N}(a;\mathcal{R}_\xi(z), I)$ be the probabilistic inverse map associated with the attribute encoder $f_\Phi$, where $z = f_\Phi(a)$ denotes the attribute embedding. The mutual information between the attribute $a$ and the attribute embedding $z$ is defined as:
		\begin{equation}
			\small
			I(a; z) = H\left(z;\Phi\right) \geq H\left(a\right) + \mathbb{E}_{a \sim p(a)} \left[\log{t_\xi(a|f_\Phi(a))}\right].    
		\end{equation}
\end{lemma}}
\noindent \textbf{Proof.} Consider the attribute $a \in \mathbb{R}^D$ and the embedding vector $z \in \mathbb{R}^d$ to be random variables under $p_\Phi(a, z) = p(a)\,p_\Phi(z|a)$ as the joint distribution. The mutual information between attribute and  $I(a; z) = H(z;\Phi) - H(z|a) = H(a) - H(a|z;\Phi)$.  As $f_\Phi : \mathbb{R}^D \rightarrow \mathbb{R}^d$ is a deterministic mapping, $p_\Phi(z|a)$ is a deterministic function of $a$, $i.e.$ $p_\Phi(z|a) = \delta(z - f_\Phi(a))$. Hence, the conditional entropy $H(z|a) = 0$, and $H(z;\Phi) = H(a) - H(a|z;\Phi)$.

\begin{align}
	\small
	H(a|z; \Phi) =& -\mathbb{E}_{p_\Phi(a,z)}\log{p_\Phi(a|z)} \nonumber\\
	= & -\mathbb{E}_{p_\Phi(a,z)}\log{t_\xi(a|z)} - \mathbb{E}_{p_\Phi(a,c)}\log{\frac{p_\Phi(a|z)}{t_\xi(a|z)}} \nonumber \\
	=& -\mathbb{E}_{p_\Phi(a,z)}\log{t_\psi(a|z)} \nonumber \\ &-\mathbb{E}_{p(z)}\left[ KL\left[{p_\Phi(a|z), t_\xi(a|z)}\right] \right] \nonumber\\
	\leq& -\mathbb{E}_{p_\Phi(a,z)}\log{t_\xi(a|z)}
\end{align}
This inequality can be used to bound the entropy:
\begin{align}
	H(z; \Phi) =& \,H(a) - H(a|z; \Phi) \\
	\geq& \, H(a) + \mathbb{E}_{p_\Phi(a,z)}\log{t_\xi(a|z)}
\end{align}
\section{Datasets}
\label{sec:apx-dataset}
We conduct experiments on five widely used datasets for zero-shot learning. CUB-200~\cite{CUB} is a fine-grain dataset containing 200 classes of birds, and AWA1~\cite{AWA1} and AWA2~\cite{xian2018zero} are datasets containing 50 classes of animals, each represented by an $85$-dimensional attribute. aPY~\cite{aPY} is a diverse dataset containing 32 classes, each associated with a 64-dimensional attribute. SUN~\cite{patterson2012sun} includes $717$ classes, each with only 20 samples; fewer samples and a high number of classes make SUN especially challenging. In the SUN dataset, each class is represented by a 102-dimensional attribute vector. The train/test split details are given in Table~\ref{tab:dataset} and the same split is used for the generalized zero-shot Learning (GZSL) setting.

The pre-processed dataset is provided by \cite{xian2018zero} and publicly available of the download\footnote{\href{1}{http://datasets.d2.mpi-inf.mpg.de/xian/xlsa17.zip}}. The dataset use ResNet-101 architecture pretrained on the ImageNet~\cite{imagenet2015} for the feature extraction of the visual domain. The features are directly extracted from the pretrained model without any finetuning. Also, the seen and unseen split proposed by \cite{xian2018zero} ensures that unseen classes are not present in the ImageNet dataset; otherwise, the zero-shot learning setting will be violated. 

\begin{table*}[ht]
	\centering
	\caption{The dataset and their split for the seen and unseen classes for the GZSL setting.}
	\addtolength{\tabcolsep}{4pt}
	\label{tab:dataset}
	\begin{tabular}{l|p{5em}|p{5em}|p{5em}|p{6em}|p{5em}|p{4em}}
		\hline
		Dataset& {Seen Classes}&Train&Val& Unseen Classes (Test)& Attribute Dimension& Total Classes  \\
		\hline
		AWA1~\cite{AWA1}& 40&30&10&10&85&50\\
		AWA2~\cite{xian2018zero}& 40&30&10&10&85&50\\
		CUB~\cite{CUB}& 150&100&50&50&312&200\\
		SUN~\cite{patterson2012sun}& 645&500&145&72&102&717\\
		aPY~\cite{aPY}& 20&15&5&12&64&32\\
		\hline
	\end{tabular}
	\vspace{1em}
\end{table*}

\section{Training and Evaluation Protocols}
In the training, first, we divide the training classes into train and validation sets as mentioned in Table-\ref{tab:dataset}. We tune the hyperparameter for the validation set that is discussed below. Once we have the optimal hyperparameter for the validation set we merge the train and validation set and retrain the model with the tuned hyperparameter and the model is evaluated for the test samples. The hyperparameters are tuned for the Generalized Zero-Shot Learning (GZSL) only and same  parameters are used for all the other experiments like Zero-shot Learning,  Fixed Continual GZSL, and  Dynamic Continual GZSL. 

We have three hyperparameters: $\lambda$, $\eta$, and $\epsilon$.
We search loss weight $\lambda$ in the interval $[0.5, 10]$ with step size $0.5$.
Learning rate $\eta$ is swept from $10^{-6}$ to $10^{-1}$ by a factor of 10, with learning rate decay with each epoch. 
We search Reptile learning rate $\epsilon$ between [$10^{-4}$, $10^{-1}$]. The final obtained hyperparameter are given in Section~\ref{sec:implementation}.
Note our baseline results are reported from \cite{gautam2021generative,gautam2020generalized,skorokhodov2021class,kuchibhotla2022unseen}; we follow the same settings and split. 

\label{sec:training-evaluation-protocol}
\subsection{Generalized Zero-Shot Learning (GZSL)}
The simplest case we consider is the generalized zero-shot learning (GZSL) setting~\cite{xian2018zero}.
In GZSL, classes are split into two groups: classes whose data are available during the model's training stage (``seen'' classes), and classes whose data only appear during inference (``unseen'' classes).
For both types, attribute vectors describing each class
are available to facilitate knowledge transfer.
During test time, samples may come from either class seen during training or new unseen classes.
We report mean seen accuracy ($mSA$) and mean unseen accuracy ($mUA$), as well as the harmonic mean ($mhM$) of both as an overall metric; harmonic mean is considered preferable to simple arithmetic mean as an overall metric, as it prevents either term from dominating~\cite{xian2018zero}. The harmonic mean ($mhM$) can be defined as:
\begin{equation}
	mhM=\frac{2\times mUA\times mSA }{ mUA+ mSA}
\end{equation} 

Note that some GZSL approaches (notably, generative ones) assume that the list of unseen classes and their attribute vectors are available during the training stage, even if their data are not; this inherently restricts these models to these known unseen classes.
Conversely, our approach only requires the attributes of the seen classes.
Also, in contrast to the continual GZSL settings described below, all seen classes are assumed available simultaneously during training.

\begin{figure*}[ht]
	\centering
	\includegraphics[width=0.47\textwidth]{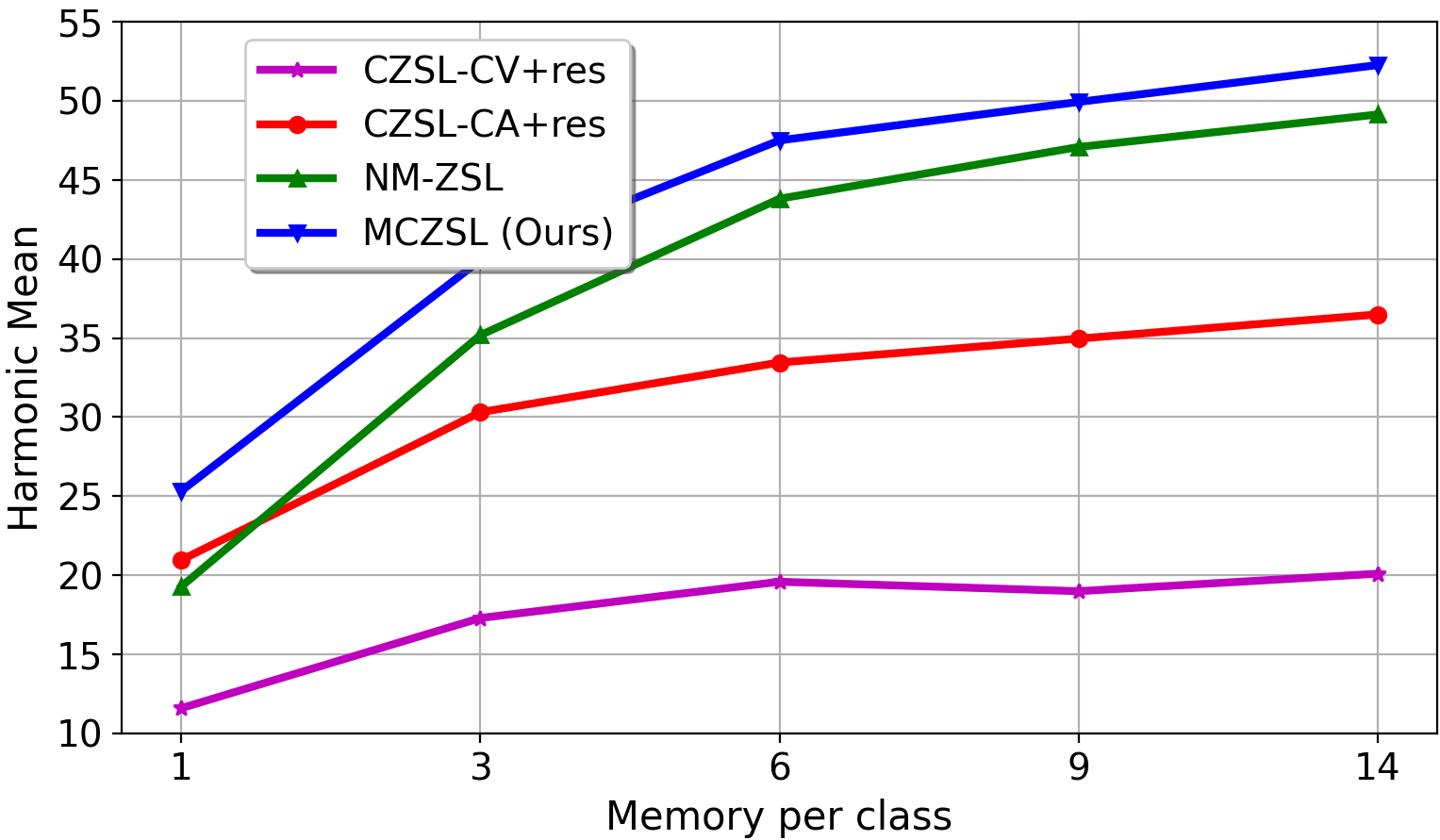} \qquad
	\includegraphics[width=0.47\textwidth]{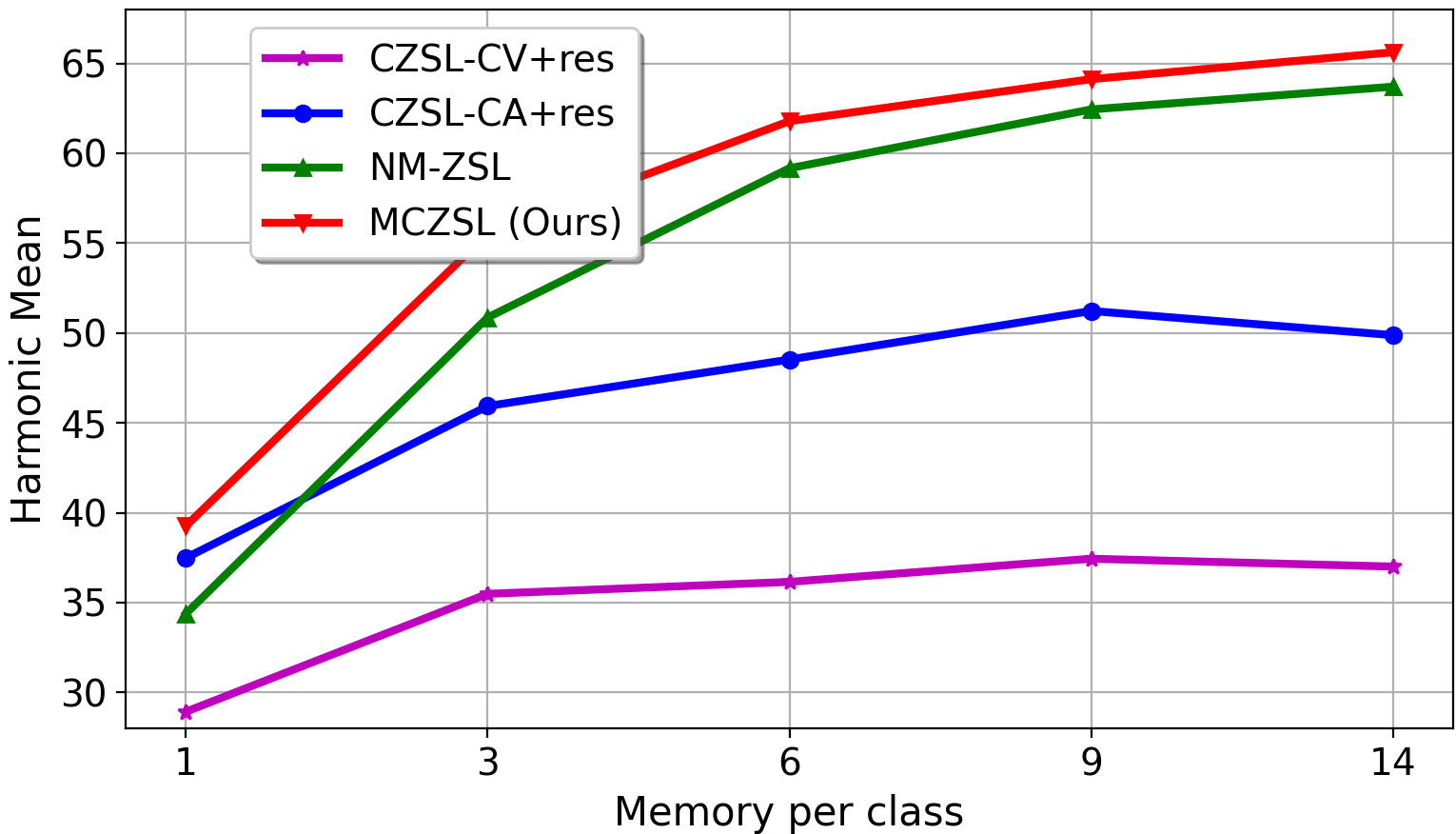}
	\caption{Model performance vs. Memory growth for continual GZSL on the CUB dataset, \textbf{Left:} Represents the Fixed Continual GZSL setting described in Section~\ref{sec:setting-1}, \textbf{Right:} Represents the Dynamic Continual GZSL setting described in Section~\ref{sec:setting2}.}
	\label{fig:memoryvs_hmean}
\end{figure*}

\subsubsection{Implementation Details}
\label{sec:implementation}
In the proposed model, $\Phi_a$, $\Phi_s$, and $\Phi_b$ are single-layer fully connected (fc) neural networks with ReLU, Sigmoid, and ReLU activation functions respectively. The dimension of each neural network ($\Phi_a$, $\Phi_s$, and $\Phi_b$ ) is $2048$. The self-gating output on the given attribute goes to another one-layer neural network of dimension $2048\rightarrow 2048$ along with the BatchNorm layer. The output of this layer is considered the projected visual feature, and in the visual space, we measure the similarity by cosine distance. For all the datasets, the model is trained for the 200 epoch per task. For the inner loop, we use Adam~\cite{kingma2014adam} optimizer with a constant learning rate $0.0001$. In the meta update, we use Adam optimizer with an initial learning rate of $0.001$, and it decreases with the increase of the epoch at a rate of $(1- current\_epoch/(total\_epoch-1))$. We follow the same hyperparameter for all the datasets that shows the model's stability and applicability for the wide range of diverse datasets. The regressor network is also a $2048$ dimensional fully connected layer, and MMR uses $\lambda=5.0$.

\subsection{Fixed Continual GZSL}
\label{sec:setting-1}

The setting proposed by \cite{skorokhodov2021class} divides all classes of the dataset into $K$ subsets, each corresponding to a task. 
For task $T_t$, the first $t$ of these subsets are considered the seen classes, while the rest are unseen; this results in the number of seen classes increasing with $t$ while the number of unseen classes decreases.
Over the span of $t=1,...K$, this simulates a scenario where we eventually ``collect'' labeled data for previously unseen classes.
Note that in contrast to the typical GZSL setting, only data from the $t^\mathrm{th}$ subset are available; previous training data are assumed inaccessible.
The goal is to learn from this newly ``collected'' data without experiencing catastrophic forgetting.
As in GZSL, we report mSA, mUA, and mH, but at the end of $K-1$ tasks: 
\begin{eqnarray}
	mSA_F=&\frac{1}{K-1}\sum_{i=1}^{K-1}\mathrm{Acc}(\cD_{ts}^i(c_{\leq i}^s),\cA(c_{\leq i}^s))\\
	mUA_F=&\frac{1}{K-1}\sum_{i=1}^{K-1}\mathrm{Acc}(\cD_{ts}^i(c_{i}^u),\cA(c_{i}^u))\\
	mhM_F=&\frac{1}{K-1}\sum_{i=1}^{K-1}\cH(\cD_{ts}^i(c_{\leq i}^s),\cD_{ts}^i(c_{i}^u),\cA)
\end{eqnarray}
where $\mathrm{Acc}$ represents per class accuracy, $\cD_{ts}^i(c_{\leq i}^s)$ and $\cA(c_{\leq t}^s)$ are the seen class test data and attribute vectors respectively during the $i^{\mathrm{th}}$ task. Similarly $\cD_{ts}^i(c_{i}^u)$ and $\cA(c_{i}^u)$ represents the unseen class test data and attribute vectors during the $i^{\mathrm{th}}$ task. $\cH$ is the harmonic mean of the accuracies obtained on $\cD_{ts}^i(c_{\leq i}^s)$ and $\cD_{ts}^i(c_{i}^u)$. We calculate the metric up to task $K-1$, as there are no unseen classes for task $K$, resulting in standard supervised continual learning.

\subsection{Dynamic Continual GZSL}
\label{sec:setting2}
While it's not unreasonable that previously unseen class may become seen in the future, the above fixed continual GZSL evaluation protocol assumes that all unseen classes and attributes are set from the beginning, which may be unrealistic.
An alternative framing of continual GZSL is one in which each task consists of its own disjoint set of seen and unseen classes, as proposed by \cite{gautam2020generalized}.
Such a formulation does not require all attributes to be known \textit{a priori}, allowing the model to continue accommodating an unbounded number of classes.
As such, in contrast to the fixed continual GZSL, the number of seen and unseen classes both increase with $t$.
As with the other settings, we report mSA, mUA and mH:

\begin{eqnarray}
	mSA_D=&\frac{1}{K}\sum_{i=1}^{K}\mathrm{Acc}(\cD_{ts}^i(c_{\leq i}^s),\cA(c_{\leq i}^s))\\
	mUA_D=&\frac{1}{K}\sum_{i=1}^{K}\mathrm{Acc}(\cD_{ts}^i(c_{\leq i}^u),\cA(c_{\leq i}^u))\\
	mhM_D=&\frac{1}{K}\sum_{i=1}^{K}\cH(\cD_{ts}^i(c_{\leq i}^s),\cD_{ts}^i(c_{\leq i}^u),\cA)
\end{eqnarray}
where $\mathrm{Acc}$ represents per class accuracy, $\cD_{ts}^i(c_{\leq i}^s)$ and $\cA(c_{\leq t}^s)$ are the seen class test data and attribute vectors during $i^{\mathrm{th}}$ task. Similarly $\cD_{ts}^i(c_{\leq i}^u)$ and $\cA(c_{\leq i}^u)$ represents the unseen class test data and attribute vector during the $i^{\mathrm{th}}$ task. Detailed splits of the seen and unseen class samples for each task are given in the supplementary material.

\subsubsection{Task Details}
The AWA1 and AWA2 datasets contain 50 classes, which are divided into five tasks of ten classes each. We divide 717 classes of the SUN dataset into 15 tasks; the first three tasks contain 47 classes, and the remainder with 48 classes each. The CUB dataset contains 200 classes; we divide all classes into 20 tasks of ten classes each. The aPY dataset contains 32 classes; we divide the dataset into four tasks with eight classes in each task. The reservoir sample $B$ for the AWA1, AWA2, CUB, SUN and aPY are $25, 25, 10, 5$ and $25$ respectively.

\begin{figure}[h]
	\centering
	\includegraphics[scale=0.22]{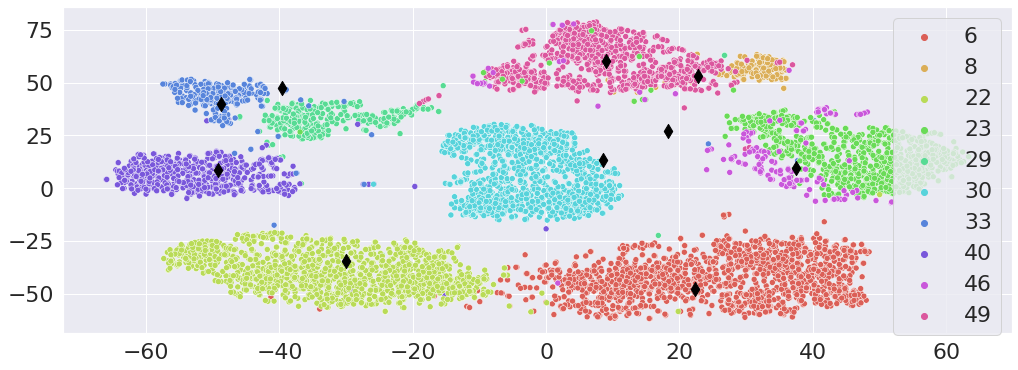}
	\caption{t-SNE plot for the unseen classes AWA2 dataset }
	\label{fig:tsne}
	\vspace{-3mm}
\end{figure}

\section{Ablation Studies}
\label{sec:ablations}
We conduct extensive ablation studies on the proposed model's different components, observing that each of the proposed components play a critical role. We show the effects of different components on the AWA1 and CUB datasets in the fixed continual GZSL setting, with more ablation studies for dynamic continual GZSL in the supplementary material. 

\subsection{Reservoir size vs Performance}
\label{sec:reservoir}
To overcome catastrophic forgetting, the model uses a constant-size reservoir~\cite{lopez2017gradient} to store previous task samples; with more tasks, the number of samples per class decreases. The reservoir size plays a crucial role in model performance. Figure~\ref{fig:memoryvs_hmean}, we evaluate the model's performance for both fixed and dynamic continual GZSL. We observe that for different reservoir sizes $\{1,3,6,9,14\} \times \# classes$, the proposed model shows consistently better results than recent models. For fixed and dynamic continual GZSL, $\# classes$ is $S+U$ and $S$, respectively. 

\subsection{t-SNE Visualization}

Figure~\ref{fig:tsne} shows the t-SNE plot for the AWA2 dataset. The attributes are projected to the visual space, and in the visual space, we do the t-SNE plot for all the samples. Here we observe that the projected attribute features closely correspond to the visual data space.

\subsection{Incorporating Generative model in the proposed approach}
As we know that the generative model shows a promising result for the GZSL setting. Here a question arises if we combined the generative and discriminative approaches, how will the model behave? To answer the above question, we perform the experiment where the generated samples of the unseen classes are also incorporated into the model during training. We observe that doing so would result in similar disadvantages as our generative baselines: slower training and reduced flexibility. The result is shown in Table~\ref{tab:gen_res_S1} where we use MZSL~\cite{verma2019meta} model to generate the unseen class samples. We observe that we still surpass all baselines with generated samples, but it does not help either. We suspect this is due to a mismatch between real and generated samples.

\begin{table}[t]
	\small
	\addtolength{\tabcolsep}{-1.2pt}
	\centering
	\caption{GZSL result when incorporating the with generated samples from MZSL~\cite{verma2019meta}}
	\label{tab:gen_res_S1}
	\begin{tabular}{l|ccc|ccc}
		\toprule
		& \multicolumn{3}{c|}{AWA1} & \multicolumn{3}{c}{CUB}  \\
		\cline{2-7}          &
		\multicolumn{1}{c}{mSA} & \multicolumn{1}{c}{mUA} & \multicolumn{1}{c|}{mH} & \multicolumn{1}{c}{mSA} & \multicolumn{1}{c}{mUA} & \multicolumn{1}{c}{mH} \\
		\midrule
		MAIN &  77.9&71.9& 74.8 & 58.7& 65.9& 62.1\\
		MAIN+MZSL [b] & 75.3& 70.1 &72.6& 57.6 & 61.7&59.5\\
		
		\bottomrule
	\end{tabular}
	\vspace{-2em}
\end{table}

\end{document}